\PassOptionsToPackage{table}{xcolor}
\documentclass{article} 
\usepackage[T1]{fontenc} 
\usepackage{iclr2027_conference,times}

\usepackage{amsmath,amsfonts,bm}

\def\eqref#1{equation~\ref{#1}}

\def\1{\bm{1}}

\DeclareMathAlphabet{\mathsfit}{\encodingdefault}{\sfdefault}{m}{sl}
\SetMathAlphabet{\mathsfit}{bold}{\encodingdefault}{\sfdefault}{bx}{n}

\DeclareMathOperator{\sign}{sign}

\renewcommand{\eqref}[1]{(\ref{#1})}

\usepackage{hyperref}
\usepackage{url}
\usepackage{xcolor}
\usepackage{listings}
\usepackage{amsmath,amssymb}
\usepackage{amsthm}
\usepackage{graphicx}
\usepackage{subcaption}
\usepackage{fontawesome5}
\usepackage{multirow}
\usepackage{booktabs}
\usepackage{wrapfig}
\usepackage{array}
\usepackage{algorithm}
\usepackage{algpseudocode}
\usepackage{placeins}
\usepackage{float}

\definecolor{oursrow}{RGB}{220,238,220}
\definecolor{projectlink}{RGB}{180,10,10}
\newcommand{\yes}{\ensuremath{\checkmark}}
\newcommand{\no}{\ensuremath{\times}}
\newcommand{\hardg}{\textbf{Hard}}
\newcommand{\up}{$\uparrow$}
\newcommand{\dn}{$\downarrow$}

\newtheorem{proposition}{Proposition}
\newtheorem{lemma}{Lemma}[section]
\DeclareMathOperator{\clip}{clip}

\title{Closed-form Action Projection for Constrained World Model Planning in Robotics}

\newcommand{\authorblock}[3]{%
    \parbox[t]{0.30\textwidth}{%
        \normalfont\raggedright
        \textbf{#1}\\
        \small #2\\
        {\fontsize{8}{10}\selectfont\texttt{#3}}
    }%
}

\author{
\authorblock
    {Hasan Taha Bagci}
    {Istanbul Technical University}
    {bagcih21@itu.edu.tr}
\And
\authorblock
    {Jieyi Bi\thanks{Corresponding author}}
    {Nanyang Technological University}
    {jieyi001@e.ntu.edu.sg}
\And
\authorblock
    {Fatma Elif Yalcindag}
    {Istanbul Technical University}
    {yalcindag22@itu.edu.tr}
\AND
\authorblock
    {Xiang Zhou}
    {XYZ Embodied AI}
    {zhoux@lamda.nju.edu.cn}
\And
\authorblock
    {Emre Koyuncu}
    {Istanbul Technical University}
    {emre.koyuncu@itu.edu.tr}
\And
\authorblock
    {Zhiguang Cao}
    {Singapore Management University}
    {zgcao@smu.edu.sg}
}

\iclrfinalcopy 
\begin{document}
\addtocontents{toc}{\protect\setcounter{tocdepth}{-1}}

\newcommand{\modelnametext}{CAP}
\DeclareRobustCommand{\modelname}[1][]{%
  \texorpdfstring{\mbox{\textsc{\modelnametext#1}\kern0.035em\raisebox{0.75ex}{\scalebox{0.62}{\faIcon{shield-alt}}}}}{\modelnametext#1}}
\maketitle
\lhead{}

\begin{abstract}
Physical control systems face hard constraints at every manipulation step, such as strict bounds on contact force and velocity. However, existing reward-driven reinforcement learning struggles to enforce these limits. This challenge is compounded in practice because constraint satisfaction depends not only on the high-level predicted action, but also on the low-level controller's live state during execution.
To bridge this gap, we introduce \textbf{\modelname{}} (Closed-form Action Projection), a modular layer that imposes hard, state-dependent action constraints on learned policies and world-model planners. At each control step, \textbf{\modelname{}} derives an admissible action set in closed form from the live controller state and projects the proposed action onto it without any online optimization or a learned constraint model, provably guaranteeing constraint satisfaction.
To resolve the geometric ambiguity inherent in pure proprioception, we integrate a world model that processes environmental observations, such as wrist depth, during impedance control. Evaluations on three representative contact-rich manufacturing tasks, peg insertion, gear meshing and nut threading, demonstrate that \textbf{\modelname{}} strictly prevents force limit violations at every step while maintaining high task success rates, achieving significant improvement against the state-of-the-art neural baselines.
\end{abstract}

\begin{center}
\textbf{Videos, models, and code:} \href{https://cap-wm.github.io}{\textcolor{projectlink}{\textbf{cap-wm.github.io}}}
\end{center}

\section{Introduction}
\label{sec:intro}
Inserting a peg, meshing gears, and threading a nut all require a robot to make contact before the task is complete. Small errors in alignment change which motions make progress and how much force those motions produce. Learning to assemble parts therefore requires both information about the surrounding geometry and control over the physical interaction~\citep{tang2023industreal,noseworthy2025forge}.

Reinforcement learning has produced effective assembly policies under pose and dynamics uncertainty~\citep{tang2024automate,noseworthy2025forge}. World models offer an additional capability by predicting the consequences of candidate actions and using those predictions to plan~\citep{wu2023daydreamer,hansen2024tdmpc2}. With wrist-depth observations, these models can also use geometric information that is difficult to recover from proprioception alone~\citep{hansen2026insertion}. This combination is attractive for assembly, where the robot must choose useful motions while working within a limited force budget.

A planner that predicts task progress still needs a way to enforce that budget during execution. In an impedance-controlled robot, the policy commands a target pose. The resulting wrench depends on controller gains, current motion, the previous filtered action, and actuation dead zones. Consequently, an action that is admissible in one controller state may produce too much force in another. A fixed action clamp cannot capture this state dependence. Reward penalties and cumulative-cost constraints encourage desirable interaction, but allow a positive force excess to be traded against progress or an episode-level budget~\citep{achiam2017cpo,stooke2020pid,huang2024safedreamer}.

We introduce \textbf{Closed-form Action Projection (\modelname)}, which converts the controller's current force budget into admissible action intervals. For the impedance controller studied here, the action-to-wrench map is affine on its unclipped branch and the actuation dead zone is monotone. These properties let us combine force, approach-speed, and target-clip constraints with coordinate-wise clipping. The resulting layer requires neither an online optimization solver nor a learned constraint model. It can act on a world-model planner or a model-free policy, giving the two learners the same form of execution constraint.

\modelname[-WM] couples depth-aware latent planning to projected execution throughout learning. The world model predicts the outcome of a proposal after CAP acts on it, allowing planning to account for the controller's correction through learned transitions. This coupling makes geometric perception, task prediction, and analytic constraint enforcement complementary. With the same execution layer at evaluation, training through CAP raises force-compliant success by 37.8 percentage points on gear meshing and 42.3 on nut threading.

Our contributions are threefold. First, we derive a closed-form projection from the live impedance-controller state and establish a bound on the modeled applied translational wrench when the admissible set is nonempty. Second, we integrate this projection with a depth-aware world model using consistent action credit during learning, and evaluate the same principle with PPO. Third, we evaluate task success and force compliance on three assembly tasks and use three-seed ablations to establish the roles of training-time projection, approach-speed constraints, and controller information. Across three training seeds, CAP-WM achieves $97.6\pm1.0\%$, $92.4\pm0.3\%$, and $89.9\pm3.1\%$ force-compliant success at 5\,N on peg insertion, gear meshing, and nut threading. These empirical rates use the filtered decision-rate wrist-force signal defined in Section~\ref{sec:problem}.

\begin{figure}[t]
\centering
\includegraphics[width=\linewidth]{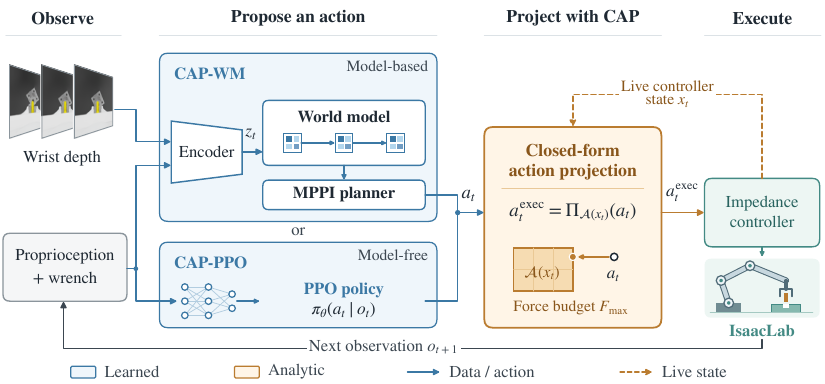}
\caption{\textbf{Two learners, one controller-aware execution layer.} CAP-WM uses wrist depth, proprioception, and wrench observations to plan with a latent world model. CAP-PPO uses a recurrent policy over proprioception and wrench observations. Both pass the proposal $a_t$ through CAP before impedance control. Blue regions denote learned components and amber denotes the analytic projection. Solid arrows carry data or actions, and dashed arrows carry the live controller state $x_t$. The camera thumbnails show simulator scenes for context. CAP-WM receives depth images. Planning and projection run at 15\,Hz and low-level control at 120\,Hz.}
\label{fig:arch}
\end{figure}

\section{Related Work}
\label{sec:related}
\textbf{Learning assembly and visual control.}
IndustReal and AutoMate learn assembly skills that transfer from simulation to hardware~\citep{tang2023industreal,tang2024automate}. FORGE adds force feedback, force-threshold conditioning, and dynamics randomization to learn robust contact behavior~\citep{noseworthy2025forge}. Visual policy learning, including Diffusion Policy~\citep{chi2023diffusion}, demonstrates the value of visual conditioning and temporally coordinated actions. World-model methods learn action-conditioned predictions for control, using latent imagination or online planning~\citep{wu2023daydreamer,hafner2025dreamerv3,hansen2024tdmpc2}. Depth-aware insertion extends this approach to geometric variation~\citep{hansen2026insertion}. CAP-WM builds on these predictive and perceptual capabilities and adds a controller-derived execution constraint.

\textbf{Constraint handling in reinforcement learning.}
CPO, PID-Lagrangian methods, and FOCOPS optimize return subject to expected cumulative costs~\citep{achiam2017cpo,stooke2020pid,zhang2020focops}. SafeDreamer combines world-model predictions with constrained planning and learning~\citep{huang2024safedreamer}. State-wise methods, shielding, and constrained action representations address requirements at individual decisions~\citep{alshiekh2018shielding,liu2022atacom,zhao2024scpo,krasowski2023provably}. These approaches make different assumptions about dynamics and constraints. Our setting uses a known controller map to express a per-decision wrench constraint, while retaining a learned model for task prediction.

\textbf{Force control and action projection.}
Model predictive force control can incorporate learned contact models and explicit uncertainty bounds~\citep{matschek2023force}. Guarding Force uses a force-constrained barrier function and quadratic programming~\citep{wang2025guarding}, while SHaRe-RL combines structured learning with per-axis compliance~\citep{stranghoener2026share}. Closed-form action corrections also appear in safety layers and hard-constrained neural networks~\citep{dalal2018safe,min2025hardnet}. CAP specializes this principle to the live impedance-controller map, including filter memory and dead zones. Its axis-wise structure makes the admissible set a box. Our experiments study both the resulting force compliance and how a learner adapts to execution through that box. Table~\ref{tab:properties} compares the constraint formulations.

\begin{table}[t]
\centering
\caption{\textbf{Capability matrix.} Each row describes the method's own constraint formulation. Closed form denotes analytic constraint enforcement without an online solve. Learned model denotes dynamics or force prediction. A dash indicates that closed-form execution filtering does not apply. Proposition~\ref{prop:force} states the CAP-WM bound; Appendix~\ref{app:calibration} specifies its calibration.}
\label{tab:properties}
\footnotesize
\setlength{\tabcolsep}{4pt}
\begin{tabular}{@{}lllcc@{}}
\toprule
\multirow{2}{*}{\textbf{Method}} & \multirow{2}{*}{\textbf{Bounded quantity}} &
\multirow{2}{*}{\textbf{Enforcement}} & \textbf{Closed} & \textbf{Learned} \\
& & & \textbf{form} & \textbf{model} \\
\midrule
FORGE \citep{noseworthy2025forge} & contact force (N) & Penalty & --- & \no \\
CPO \citep{achiam2017cpo} & cumulative cost & Budget & --- & \no \\
SafeDreamer \citep{huang2024safedreamer} & cumulative cost & Budget & \no & \yes \\
SHaRe-RL \citep{stranghoener2026share} & interaction force (N) & Compliance & \yes & \no \\
\citet{matschek2023force} & contact force (N) & Chance & \no & \yes \\
\rowcolor{oursrow}
\textbf{\modelname[-WM] (ours)} & applied wrench (N) & \hardg{} & \yes & \yes \\
\bottomrule
\end{tabular}
\end{table}

\section{Problem Setting}
\label{sec:prelim}\label{sec:problem}\label{sec:controller}
We consider assembly as an episodic, partially observed decision process. At step $t$, the robot observes $o_t$, proposes $a_t\in\mathcal A=[-1,1]^n$, and receives reward $R(s_t,a_t^{\mathrm{exec}})$ after the executed action changes the state. The goal is high expected return under a force requirement
\begin{equation}
\max_\pi\;J(\pi)=\mathbb E_\pi\!\left[\sum_t\gamma^t R(s_t,a_t^{\mathrm{exec}})\right],
\qquad \|F_t\|_2\le F_{\max}\quad\text{at each decision step}.
\label{eq:problem}
\end{equation}
Throughout this paper, measured force denotes the three-dimensional wrist force supplied by FORGE's force-sensing interface. We use its native low-pass filtered signal, sampled at the end of each executed control period before observation noise is added. Thus $F_t$ in Eq.~\eqref{eq:problem} specifies the wrist-force requirement at each decision step. Section~\ref{sec:admissible} connects this signal to the controller-output bound, and Appendix~\ref{app:observations} specifies the filter and sampling convention.

The agent updates the target at 15\,Hz. The impedance controller holds that target for eight 120\,Hz physics steps and recomputes its wrench from the changing robot state. This hierarchy allows planning and low-level execution to run at different rates. Each action contains translation, rotation, and a success prediction. FORGE fixes roll and pitch, leaving translation and yaw as the physical command channels. We retain its task rewards and success criteria (Appendix~\ref{app:rewards}).

The action is first smoothed, $\tilde a_t=\lambda a_t+(1-\lambda)\tilde a_{t-1}$. Its translational channels set the target $p_t^{\mathrm d}=p^{\mathrm{ref}}+B\tilde a_{t,1:3}$. Stiffness $K_p$ converts target error into a wrench and damping $K_d$ opposes motion. A position-error clip limits the controller's operating range, and a dead zone absorbs small commanded wrenches. Episode randomization changes these quantities. The learned agent receives sensor observations and history, while CAP reads the controller interface directly. This distinction allows perception and task prediction to remain learned while the execution constraint uses the implemented control law. Appendix~\ref{app:impedance} gives the complete controller equations.

\section{Method}
\label{sec:method}
CAP-WM combines a depth-aware planner, a controller-derived projection, and learning from their joint execution (Figure~\ref{fig:arch}). We first describe the planner, then the admissible set, and finally how the model and policy learn through the projection.

The projection specifies admissible motion, while the learner chooses proposals using task observations and reward. Coordinate-wise clipping preserves a proposal inside the box and changes only coordinates outside their intervals. Even when several coordinates reach a boundary, the policy still selects the direction and sequence of motion as the controller state evolves. Wrist depth informs this choice through geometric observations, while the analytic layer uses the current controller map.

\subsection{Depth-Aware World-Model Planning}
\label{sec:wm}\label{sec:capwm}
The observation combines wrist depth $I_t$ with proprioception, force, and recent executed commands. Depth supplies relative geometric information. A short history of controller readings supplies evidence about the randomized dynamics. Separate encoders fuse the modalities into a latent state $z_t=h_\phi(o_t)$.

We use TD-MPC2's latent dynamics $f_\phi$, reward predictor $R_\phi$, value ensemble $Q_\phi$, and policy prior $\pi_\phi$~\citep{hansen2024tdmpc2}. The planner rolls out candidate action sequences in latent space, scores their predicted rewards and terminal values, and executes the first action of the selected sequence after CAP projection. The base planning horizon is three decisions, with value prediction accounting for later returns. The planning objective is
\begin{equation}
\begin{aligned}
\max_{a_{t:t+H-1}\in\mathcal{A}^H}
&\sum_{k=0}^{H-1}\gamma^k
R_\phi(z_{t+k},a_{t+k})
+\gamma^H Q_\phi(z_{t+H},a_{t+H}^{\pi}) \quad
\text{s.t.} \ \ z_{t+k+1}=f_\phi(z_{t+k},a_{t+k}),
\end{aligned}
\label{eq:mpc}
\end{equation}
where $a_{t+H}^{\pi}$ is drawn from the learned policy prior. The world model is trained from online interaction. Depth frames are observations, and its supervision comes from collected transitions, rewards, and value targets.

We use a $96\times96$ depth image and histories of 3 decisions for peg insertion and 32 for gear meshing and nut threading. The representation and planner follow the existing depth-aware TD-MPC2 design. The integration with CAP concerns the execution map and the transitions used for learning. Appendix~\ref{app:hyper} lists the network and planning hyperparameters.

The depth encoder is a four-layer strided convolutional network. Its output is added to the encoded controller history and passed through a fusion network. The history includes pose, velocity, wrist force, and executed actions, so it carries evidence about how recent commands moved the robot. MPPI~\citep{williams2018mppi} samples candidate sequences, predicts their returns, and refits its action distribution to high-scoring candidates. CAP acts on the selected first action using the current controller state. Training the dynamics on projected execution lets these latent rollouts predict the consequences of that proposal convention.

\subsection{Closed-Form Action Projection}
\label{sec:admissible}

\begin{figure}[t]
\centering\includegraphics[width=\linewidth]{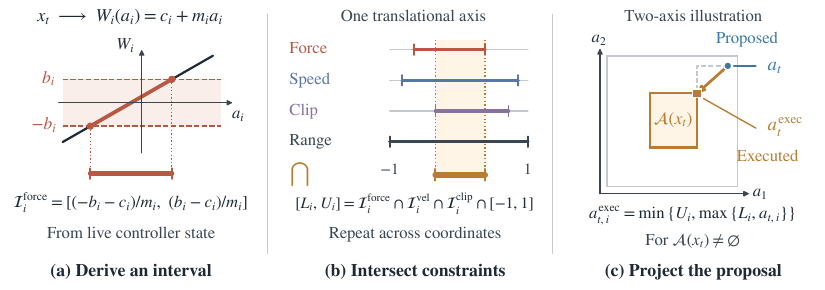}
\caption{\textbf{How physical bounds become an action correction.} (a) On the unclipped branch, the controller maps $a_i$ to $W_i=c_i+m_i a_i$. The band $[-b_i,b_i]$ includes dead-zone credit and the drift allowance from Eq.~\eqref{eq:main-force}. Its preimage is the force interval. (b) The force, speed, target-clip, and action-range intervals are intersected on each translational axis. (c) Coordinate-wise clipping projects a proposal onto the resulting nonempty box. The coefficients and geometry are illustrative. Empty intersections use the priority rule described in the text.}
\label{fig:mechanism}
\end{figure}
\textbf{From controller state to force.}
Let $x_t$ contain the gains, filter state, dead-zone widths, robot pose and velocity, and the geometric and dynamic quantities used to construct the intervals. On the branch where target-error clipping is inactive, the translational controller map on axis $i$ is
\begin{equation}
W_{t,i}(a)=c_i(x_t)+m_i(x_t)a_i,\qquad
W^{\mathrm{app}}_{t,i}(a)=D_{d_i}(W_{t,i}(a)),\quad m_i>0,
\label{eq:main-map}
\end{equation}
where $D_d(w)=\operatorname{sign}(w)\max(0,|w|-d)$. The offset $c_i$ includes the previous smoothed action and current motion. The slope $m_i$ includes the gain, action scale, and smoothing coefficient. Appendix~\ref{app:derivation} derives both directly from the controller.
The offset changes with the current target error and velocity, even if the gains remain fixed. Smoothing also makes the previous executed command relevant. These are the quantities that move the admissible interval between decisions.

\textbf{Constructing the intervals.}
A step budget $\bar F_t\le F_{\max}-\varepsilon$ is divided among the translational axes using positive weights $\kappa_i$ with $\|\kappa\|_2\le1$. Inverting the monotone controller map gives a force interval $\mathcal I_i^{\mathrm{force}}$. Writing $\eta_i=K_{p,i}|v_{t,i}|\Delta t$ for the drift allowance gives the explicit force interval
\begin{equation}
b_i=d_i+(\kappa_i\bar F_t-\eta_i)_+,\qquad
\mathcal I_i^{\mathrm{force}}=
\left[\frac{-b_i-c_i}{m_i},\frac{b_i-c_i}{m_i}\right].
\label{eq:main-force}
\end{equation}
The dead-zone width $d_i$ converts an allowed applied wrench into its commanded-wrench preimage (Appendix~\ref{app:proof-lemmas}). The drift allowance reserves part of the axis budget for motion while the target is held. A second interval $\mathcal I_i^{\mathrm{vel}}$ bounds the next velocity predicted by a local free-body model and reduces approach speed near contact. A third interval $\mathcal I_i^{\mathrm{clip}}$ keeps the controller on its affine branch. Near the mating surface, the speed budget approaches a calibrated touchdown speed. Farther away, a braking curve permits faster motion. Inverting the monotone dead-zone map converts the corresponding applied-wrench bounds into $\mathcal I_i^{\mathrm{vel}}$. The target-clip interval imposes $|c_i+K_{d,i}v_{t,i}+m_i a_i|\le K_{p,i}\delta_i$. Figure~\ref{fig:mechanism} shows how the three intervals combine. Their intersection with the action range yields
\begin{equation}
[L_i,U_i]=\mathcal I_i^{\mathrm{force}}\cap\mathcal I_i^{\mathrm{vel}}\cap
\mathcal I_i^{\mathrm{clip}}\cap[-1,1],\qquad
 a^{\mathrm{exec}}_{t,i}=\min\{U_i,\max\{L_i,a_{t,i}\}\}.
\label{eq:main-proj}
\end{equation}
For a nonempty intersection, this is the Euclidean projection onto the admissible box $\mathcal A(x_t)$. Appendix~\ref{app:derivation} gives the controller derivation and explicit endpoints in Appendix~\ref{app:proof-lemmas}. When the intersection is empty, the implementation applies target-clip, speed, and force clips in sequence, followed by the action-range clip. This priority rule always returns a bounded command. The force-only corollary requires the final action-range clip to leave the force correction unchanged and the target-error clip to remain inactive. Appendix~\ref{app:proof-practice} details these conditions. Gear and nut configurations add geometric yaw constraints described in Appendix~\ref{app:keepout}.

\begin{proposition}[Controller-output bound]
\label{prop:force}
Under the controller-map assumptions of Appendix~\ref{app:proof-setting}, if $\mathcal A(x_t)$ is nonempty, projecting any proposal onto it gives
$|W^{\mathrm{app}}_{t,i}|\le\kappa_i\bar F_t$ and
$\|W^{\mathrm{app}}_t\|_2\le\bar F_t\le F_{\max}-\varepsilon$.
The projected action also satisfies the per-axis velocity bound in the local free-body model.
\end{proposition}
The proof follows from interval membership and $\|\kappa\|_2\le1$ (Appendix~\ref{app:proof-prop}). Equal allocation, $\kappa_i=1/\sqrt3$, provides a direction-neutral default. A geometry-dependent allocation preserves this instantaneous result if the weights satisfy the same norm condition and the intervals are recomputed.

The axis budgets define a box inside the allowed wrench-norm ball. Inverting the controller map gives the corresponding box in normalized action coordinates, where projection minimizes the Euclidean change to the proposal. State-dependent offsets and slopes connect this separable correction to physical actuation.

\textbf{From the controller bound to the force requirement.}
CAP enforces the modeled-wrench budget analytically, independently of prediction accuracy or the proposed action. For a nonempty admissible set, Proposition~\ref{prop:force} gives $\|W_t^{\mathrm{app}}\|_2\le\bar F_t$. Under the sensor-to-controller condition $\|F_t\|_2\le\|W_t^{\mathrm{app}}\|_2+\varepsilon$, reserving $\varepsilon$ yields $\|F_t\|_2\le\bar F_t+\varepsilon\le F_{\max}$. Separate traces calibrate this margin for the filtered 15\,Hz wrist signal (Appendix~\ref{app:calibration}); the speed interval regulates approach motion. Corollary~\ref{cor:unsaturated} extends the wrench bound to force-priority fallback under its clip conditions, with execution diagnostics in Table~\ref{tab:diagnostics}. Appendix~\ref{app:proof-practice} states the within-period conditions.

CAP uses a controller interface providing gains, smoothing and filter memory, dead-zone widths, and robot state. The speed and yaw intervals additionally use task geometry and directional inertia. In the simulation, these are the realized controller and geometric quantities; the redraw guard uses the configured schedule. Appendices~\ref{app:state} and~\ref{app:observations} specify these inputs and the wrist-force measurement convention.

Separate calibration runs determine the measured-force margin, an impact slope relating approach speed to force excess, and a gap-dependent speed budget. The budget also decreases when the previous measured force approaches the cap. Appendix~\ref{app:calibration} details this calibration. Since FORGE redraws dead zones on a known schedule, a guard reduces dependence on dead-zone credit before each redraw (Appendix~\ref{app:guard}). These settings are fixed for evaluation. Appendix~\ref{app:implementation} gives their estimation rules, values, and the complete control loop.
For example, peg insertion at 5\,N uses $\varepsilon=0.784$\,N, leaving a 4.216\,N modeled-wrench budget. The calibrated impact allowance of 2.0\,N and slope of 201.7\,N$\cdot$s/m give a touchdown speed of about 9.9\,mm/s. This translates a task-level force requirement into an approach-speed setting with physical units.

The interval calculation uses a constant number of operations per axis. On one CPU thread, its arithmetic and diagnostics take 0.34--0.39\,ms for 128 environments. Inertia estimation additionally uses one $7\times7$ linear solve per environment. Appendix~\ref{app:state} gives the benchmark setup.

\subsection{Learning Through Projected Execution}
\label{sec:training}
The planner chooses a proposal $a_t$, while the environment receives $a_t^{\mathrm{exec}}$. We store the proposal with the transition it produces through CAP. Thus, latent dynamics and value predictions learn the composed transition
\begin{equation}
(s_t,a_t)\longmapsto P\big(\,\cdot\mid s_t,\Pi_{\mathcal A(x_t)}(a_t)\big).
\label{eq:composed}
\end{equation}
The observation history receives the executed command because that command updates the controller's filter memory. Projection is active from the random seed phase onward. This gives training and planning the same proposal-to-transition convention.

CAP-PPO uses the same execution layer with a recurrent Gaussian policy trained from scratch. PPO~\citep{schulman2017ppo} computes its likelihood ratio on the sampled proposal, whose probability the policy defines, and learns from the reward earned after projection. Both learners can therefore adapt their proposals to useful motion under the constraint. Appendix~\ref{app:action-credit} gives the corresponding update conventions.

Equation~\eqref{eq:composed} gives the planner a consistent prediction target, including when different proposals share an executed command. The executed-command history exposes the filter memory relevant to the next decision. The training-time projection ablation tests this coupling with CAP active in both evaluation arms; the approach-speed ablation tests which admissible motions enable progress under the force budget.

\section{Experiments}
\label{sec:experiments}\label{sec:setup}
We evaluate four questions. Can CAP preserve task completion while improving force compliance? Does learning with projection improve on adding it at evaluation? Which controller-dependent constraints matter? How consistent is task performance across training seeds?

\textbf{Tasks and comparisons.}
We use FORGE peg insertion, gear meshing, and nut threading~\citep{noseworthy2025forge}. Appendices~\ref{app:tasks} and~\ref{app:observations} detail the environments and observations, with full-robot and contact views in Figure~\ref{fig:simulation-scenes}. They require alignment, interaction with neighboring geometry, and coordinated translation and rotation. We compare PPO (FORGE), TD-MPC2 with depth, CAP-PPO, and CAP-WM, followed by constrained-RL and safety-layer baselines. The constrained-RL family contains CPO, PID-Lagrangian PPO, and FOCOPS. The safety-layer family contains SC$^3$ and FCSCC with several stiffness settings. Each family reports its selected member at each cap. Appendix~\ref{app:baselines} gives their implementations. Appendix~\ref{app:config} specifies observations and training recipes.

\textbf{Protocol and metrics.}
We use $F_{\max}\in\{5,10\}$\,N, the endpoints of FORGE's force-penalty threshold range. These are evaluation budgets for assembly. SR measures task completion and SSR measures completion with every recorded $\|F_t\|_2$ within the budget. Thus, an episode with low force but no completion contributes zero to SSR. EF integrates the positive force excess over each episode and averages across episodes. PF averages their peak force norms (Appendix~\ref{app:metrics}). The main comparison uses 640 episodes per cell. Training-seed tests use draw 100, with checkpoint protocols specified in Appendix~\ref{app:new-seeds}. Ablations use matched evaluation conditions and three training seeds, with 640 episodes per seed. World-model and model-free training use 32 and 128 parallel environments respectively. Appendix~\ref{app:config} gives the full protocols, with matched evaluation settings in Section~\ref{app:new-protocol}, checkpoint selection in Section~\ref{app:selection}, and compute in Section~\ref{app:compute}.

\subsection{Task Completion Under a Force Budget}
\label{sec:results}
Table~\ref{tab:main} shows the main comparison across all three tasks and both force limits. At 5\,N, CAP-WM reaches 98.1\%, 94.8\%, and 89.2\% SSR. The unprojected visual planner retains high task success but has at most 0.6\% SSR at this limit. CAP-PPO also substantially improves force-compliant completion, showing that the controller-derived layer supports both learning backbones. At 10\,N, CAP-WM leads on peg insertion and nut threading, while CAP-PPO achieves the highest SSR on gear meshing.

\begin{table}[t]
\centering\footnotesize\setlength{\tabcolsep}{3.2pt}
\caption{\textbf{Task success and force compliance.} Each result uses 640 episodes. SR and SSR are percentages, EF is mean integrated force excess in N$\cdot$s, and PF is mean episode peak force in N. The baseline-family members and checkpoint selection are detailed in Appendices~\ref{app:families} and~\ref{app:selection}. Green rows denote our methods. Bold marks the best reported value for each task, force limit, and metric, including ties.}
\label{tab:main}
\begin{tabular}{@{}>{\centering\arraybackslash}p{\dimexpr0.045\linewidth-\tabcolsep\relax}>{\raggedright\arraybackslash}p{\dimexpr0.245\linewidth-2\tabcolsep\relax}>{\raggedleft\arraybackslash}p{\dimexpr0.08875\linewidth-2\tabcolsep\relax}>{\raggedleft\arraybackslash}p{\dimexpr0.08875\linewidth-2\tabcolsep\relax}>{\raggedleft\arraybackslash}p{\dimexpr0.08875\linewidth-2\tabcolsep\relax}>{\raggedleft\arraybackslash}p{\dimexpr0.08875\linewidth-2\tabcolsep\relax}>{\raggedleft\arraybackslash}p{\dimexpr0.08875\linewidth-2\tabcolsep\relax}>{\raggedleft\arraybackslash}p{\dimexpr0.08875\linewidth-2\tabcolsep\relax}>{\raggedleft\arraybackslash}p{\dimexpr0.08875\linewidth-2\tabcolsep\relax}>{\raggedleft\arraybackslash}p{\dimexpr0.08875\linewidth-\tabcolsep\relax}@{}}
\toprule
&&\multicolumn{4}{c}{$F_{\max}=5$\,N}&\multicolumn{4}{c}{$F_{\max}=10$\,N}\\
\cmidrule(lr){3-6}\cmidrule(l){7-10}
&\textbf{Method}&\textbf{SR}\up&\textbf{SSR}\up&\textbf{EF}\dn&\textbf{PF}\dn&\textbf{SR}\up&\textbf{SSR}\up&\textbf{EF}\dn&\textbf{PF}\dn\\
&& (\%) & (\%) & (N$\cdot$s) & (N) & (\%) & (\%) & (N$\cdot$s) & (N)\\
\midrule
 & PPO (FORGE) & 97.7 & 0.3 & 2.424 & 10.26 & 97.7 & 49.5 & 0.122 & 10.26 \\
 & TD-MPC2 + depth & 99.1 & 0.6 & 7.784 & 10.60 & 99.1 & 44.7 & 0.284 & 10.60 \\
 & Safe RL & 35.9 & 25.2 & 0.034 & 4.26 & 77.8 & 77.2 & \textbf{0.000} & 4.94 \\
 & Safety layer & 90.2 & 6.1 & 0.243 & 7.01 & 91.9 & 58.3 & 0.559 & 9.58 \\
\cmidrule(l){2-10}
\rowcolor{oursrow}
\cellcolor{white} & \textbf{\modelname[-PPO]} & 78.0 & 78.0 & \textbf{0.000} & 1.75 & 90.5 & 90.5 & \textbf{0.000} & 3.45 \\
\rowcolor{oursrow}
\cellcolor{white}\multirow{-6}{*}{\rotatebox[origin=c]{90}{Peg insertion}} & \textbf{\modelname[-WM]} & 98.1 & \textbf{98.1} & \textbf{0.000} & \textbf{1.56} & 98.0 & \textbf{98.0} & \textbf{0.000} & \textbf{3.15} \\
\midrule
 & PPO (FORGE) & 99.8 & 0.0 & 12.957 & 15.10 & 99.8 & 11.6 & 2.274 & 15.10 \\
 & TD-MPC2 + depth & 95.2 & 0.0 & 20.253 & 13.48 & 95.2 & 11.6 & 1.849 & 13.48 \\
 & Safe RL & 67.2 & 8.0 & 0.252 & 7.37 & 58.3 & 50.8 & 0.018 & 7.89 \\
 & Safety layer & 21.7 & 2.2 & 0.632 & 5.01 & 87.3 & 52.7 & 0.342 & 10.14 \\
\cmidrule(l){2-10}
\rowcolor{oursrow}
\cellcolor{white} & \textbf{\modelname[-PPO]} & 94.7 & 94.7 & \textbf{0.000} & \textbf{1.89} & 98.4 & \textbf{98.4} & \textbf{0.000} & 3.05 \\
\rowcolor{oursrow}
\cellcolor{white}\multirow{-6}{*}{\rotatebox[origin=c]{90}{Gear meshing}} & \textbf{\modelname[-WM]} & 94.8 & \textbf{94.8} & \textbf{0.000} & 1.96 & 95.0 & 95.0 & \textbf{0.000} & \textbf{3.00} \\
\midrule
 & PPO (FORGE) & 97.5 & 0.0 & 36.907 & 12.45 & 97.5 & 18.6 & 5.465 & 12.45 \\
 & TD-MPC2 + depth & 96.2 & 0.0 & 43.143 & 12.72 & 96.2 & 17.7 & 3.907 & 12.72 \\
 & Safe RL & 89.5 & 0.3 & 2.239 & 8.63 & 90.6 & 68.6 & 0.039 & 9.33 \\
 & Safety layer & 45.0 & 1.1 & 0.787 & 6.66 & 77.5 & 72.0 & 0.059 & 8.49 \\
\cmidrule(l){2-10}
\rowcolor{oursrow}
\cellcolor{white} & \textbf{\modelname[-PPO]} & 61.4 & 61.4 & \textbf{0.000} & 1.85 & 87.2 & 87.2 & \textbf{0.000} & 3.22 \\
\rowcolor{oursrow}
\cellcolor{white}\multirow{-6}{*}{\rotatebox[origin=c]{90}{Nut threading}} & \textbf{\modelname[-WM]} & 89.2 & \textbf{89.2} & \textbf{0.000} & \textbf{1.76} & 91.9 & \textbf{91.9} & \textbf{0.000} & \textbf{3.18} \\
\bottomrule
\end{tabular}
\end{table}

Table~\ref{tab:main} shows that CAP combines task completion with force compliance across the three tasks and both caps. Figure~\ref{fig:seeds} complements this comparison with tests across independently trained seeds.

\begin{figure}[t]
\centering\includegraphics[width=\linewidth]{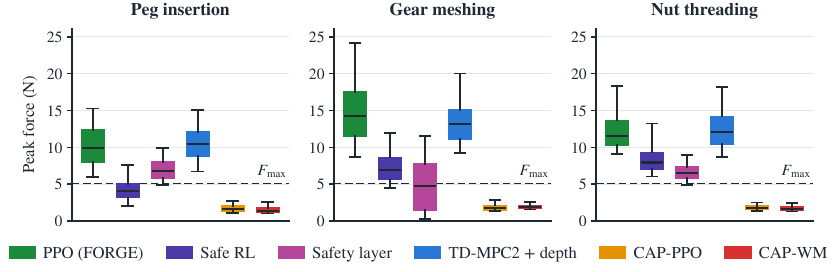}
\caption{\textbf{Episode peak force at a 5\,N cap.} Each box summarizes 640 episodes from the corresponding main-table cell. The line is the median, boxes show the 25th--75th percentiles, and whiskers show the 5th--95th percentiles. The dashed line marks the cap. Complete exceedance curves appear in Figure~\ref{fig:exceedance}. Method colours are shared across the experiment plots.}
\label{fig:analysis}
\end{figure}

Figure~\ref{fig:analysis} shows the distribution behind these aggregate rates. At 5\,N on peg insertion, CAP-WM reduces mean episode peak force from TD-MPC2's 10.60\,N to 1.56\,N. The gear and nut distributions show the same shift toward lower forces while retaining high completion. Appendix~\ref{app:additional} gives further evaluations, including full distribution tails in Section~\ref{app:force-distribution}.

\begin{figure}[t]
\centering\includegraphics[width=\linewidth]{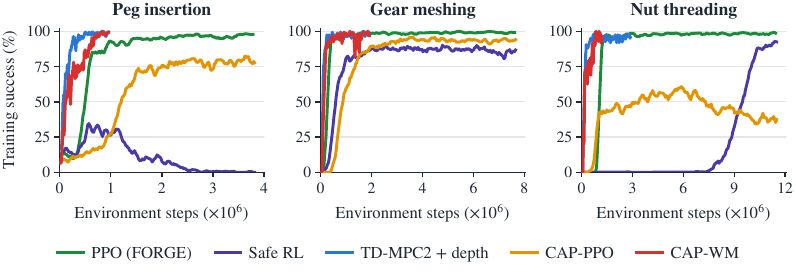}
\caption{\textbf{Training success for the 5\,N comparison.} Each curve follows one training run, with a causal moving average over approximately 2.5\% of logged points. The horizontal axis counts interactions across all environments. World-model agents use one quarter of the model-free interaction budget. Frozen safety layers have no training curve. These curves show task success during training, while Table~\ref{tab:main} reports selected-checkpoint evaluation.}
\label{fig:learning}
\end{figure}

\textbf{Learning behavior.}
Figure~\ref{fig:learning} traces the training runs behind the 5\,N cells. CAP-WM learns successful task behavior within the smaller world-model interaction budget. CAP-PPO learns reliably on gear meshing, while its peg and nut training success remains lower. The unconstrained learners also reach high training success, which makes the force measurements essential for assessing the resulting behavior. These curves describe one training seed per method. The separate seed tests in Section~\ref{sec:seeds} quantify variation between trained models.

\subsection{Why Train with CAP?}
\label{sec:train-ablation}
Adding CAP to a trained planner immediately changes how its actions affect the robot. Figure~\ref{fig:ablations}(a) tests whether training with those changes helps. In the matched-platform ablation evaluation, gear meshing with evaluation-only projection achieves $55.3\pm16.2\%$ SSR, while CAP-WM trained with projection achieves $93.1\pm0.8\%$. On nut threading, the corresponding rates are $47.6\pm7.6\%$ and $89.9\pm3.0\%$. Values are means and sample standard deviations over three training seeds. These Full-CAP reference evaluations use the same platform as the ablated arms; Figure~\ref{fig:seeds} reports the separate training-seed tests. A separate peg experiment also favors training with projection, with 98.3\% SSR versus 81.1\% for evaluation-only projection (one training seed).

The complete task-specific execution layer is active in both evaluation arms. The gains therefore show the value of learning task predictions and behavior through that layer: the planner learns which proposals produce useful progress after correction. This integration is a central contribution of CAP-WM, supported across three training seeds on both tasks. Appendix~\ref{app:component-ablations} gives the settings and complete results.

\subsection{Which Constraints Matter?}
\label{sec:components}
\begin{figure}[t]
\centering\includegraphics[width=\linewidth]{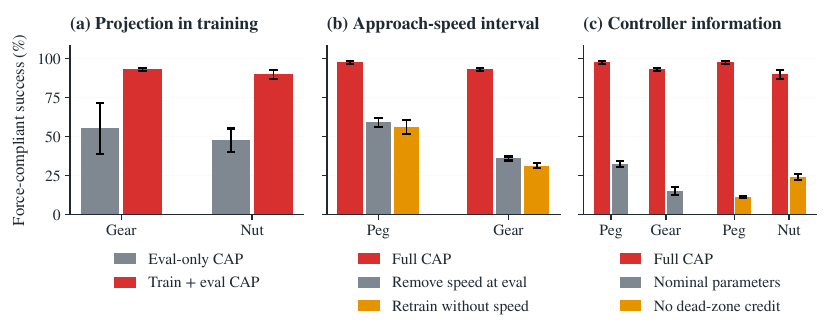}
\caption{\textbf{Learning and controller dynamics both matter.} At a 5\,N cap, (a) training with projection improves force-compliant completion over evaluation-only projection, (b) removing the approach-speed interval reduces SSR even after retraining, and (c) using nominal controller parameters or removing dead-zone credit also reduces SSR. Bars show means and error bars show sample standard deviations over three training seeds. }
\label{fig:ablations}
\end{figure}

\textbf{Approach-speed constraints.}
On peg insertion, removing the speed interval at evaluation reduces SSR from $97.6\pm1.0\%$ to $59.1\pm2.8\%$. Retraining without it yields $56.0\pm4.5\%$ despite $99.2\%$ task success. On gear meshing, the same sequence gives $93.1\pm0.8\%$, $35.9\pm1.3\%$, and $31.4\pm1.5\%$ SSR. The retrained arms retain the force and target-clip intervals, calibration, and budget feedback. Learning thus benefits from explicitly constraining the approach dynamics as well as the modeled wrench (Figure~\ref{fig:ablations}(b)).

The evaluation-only removal tests the response of a policy learned with the full execution map. Retraining tests whether the learner can adapt to the reduced constraint set over its training budget. Reading both arms together distinguishes an immediate change in execution from the performance of a policy optimized with that change. The persistent SSR gap in the retrained arms supports retaining the approach-speed interval as part of the learning environment.

\textbf{Live controller parameters and dead zones.}
Live controller information enables CAP to retain useful motion under the force budget. With live parameters, SSR reaches 97.6\% on peg insertion and 93.1\% on gear meshing, compared with 32.4\% and 15.2\% using nominal parameters. Accounting for the dead zone raises peg success from 11.1\% to 97.6\%. These fixed-policy comparisons identify the contribution of the controller-aware intervals (Figure~\ref{fig:ablations}(c); Appendix~\ref{app:component-ablations}); Appendix~\ref{app:new-audits} evaluates bounded dead-zone estimates.

\textbf{Intervention and execution changes.}
CAP modifies translation on 94.6--99.4\% of decisions in the main-table evaluations. This frequent intervention is consistent with learning a behavior that uses the constrained execution map. Removing projection from those trained controllers reduces SSR to at most 0.9\% across tasks and caps. An evaluation-time force gate on unprojected policies reduces accumulated excess force, yet at least 632 of 640 episodes still exceed the 5\,N cap. The gate acts after the previous force reading crosses its threshold. These fixed-checkpoint tests complement the retrained ablations by examining execution changes directly (Appendices~\ref{app:projection-activity} and~\ref{app:execution-ablations}).

\subsection{Visual Planning and Training-Seed Variation}
\label{sec:seeds}
\textbf{Geometric observations for constrained task progress.}
In the peg-insertion observation study, wrist depth raises projected-world-model success from 85.2\% to 98.1\%, a 12.9-point gain, with zero observed violations in both evaluations. The visual model reaches this result with 0.96M interactions, compared with 3.20M for the proprioceptive model (Table~\ref{tab:modality}). This single-seed study supports depth as a source of useful task information alongside analytic force control. Appendix~\ref{app:depth-study} gives the settings, and Table~\ref{tab:horizon} evaluates longer inference horizons using fixed checkpoints.

\begin{figure}[t]
\centering\includegraphics[width=\linewidth]{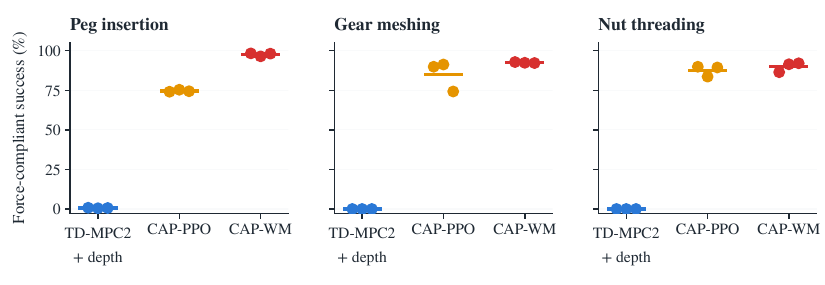}
\caption{\textbf{Force-compliant success across training seeds at 5\,N.} Each point is one seed tested on draw 100 with 640 episodes; horizontal marks indicate the mean. Training protocols and full counts are in Appendix~\ref{app:new-seeds} and Table~\ref{tab:new-seeds}.}
\label{fig:seeds}
\end{figure}

Figure~\ref{fig:seeds} reports tests across three training seeds. At 5\,N, CAP-WM obtains $97.6\pm1.0\%$ SSR on peg insertion, $92.4\pm0.3\%$ on gear meshing, and $89.9\pm3.1\%$ on nut threading. The CAP-WM nut tests and CAP-PPO peg tests each record zero force-cap violations across 1,920 episodes. CAP-PPO obtains $87.6\pm3.5\%$ SR and SSR on nut threading, with zero observed filtered-force violations in 1,920 test episodes. Across-seed summaries and training protocols appear in Appendix~\ref{app:new-seeds}.

Together, these studies support three complementary capabilities: depth-informed task progress, learning the consequences of projected execution, and high force-compliant completion across training seeds. CAP-WM combines them in a visual planning system. The strong nut CAP-PPO result additionally demonstrates that the same projection principle supports a model-free learner, extending CAP's applicability across learning backbones.

\FloatBarrier
\section{Discussion and Conclusion}
\label{sec:conclusion}
CAP turns the live state of an impedance controller into a closed-form action constraint. Combining it with depth-aware planning gives a learner both geometric information for task progress and a controller-derived limit on execution. The assembly results show high force-compliant success, and the ablations establish that training with projection and constraining approach speed each make a substantial contribution.

\textbf{Limitations and future work.}
The evaluation covers three simulated assembly tasks with task-specific geometric intervals. Hardware evaluation will require an interface or identification procedure supplying the controller quantities specified in Section~\ref{sec:admissible}, together with calibrated force sensing. The controller-information ablations and bounded-estimate study provide a basis for this extension (Appendices~\ref{app:component-ablations} and~\ref{app:new-audits}). Broader task geometries, progress-aware recovery, and recursive feasibility are further directions.

\section*{AI Use Statement}
We used large language models as writing assistants to improve the clarity and phrasing of the
text, and as search aids when surveying related work. The research questions, the method, and the
theoretical results were developed by the authors without LLM involvement, and all content was
checked.

\section*{Reproducibility Statement}
Appendices provide the controller-map derivation and proofs (Appendix~\ref{app:proof}), calibration procedure, training settings, checkpoint protocols, and evaluation metrics. The results include component ablations and training-seed variability. The project page is \url{https://cap-wm.github.io}.

\bibliography{iclr2027_conference}
\bibliographystyle{iclr2027_conference}
\clearpage\appendix
\definecolor{appblue}{RGB}{230,239,247}
\definecolor{appgreen}{RGB}{232,243,234}
\definecolor{appgray}{RGB}{245,247,249}
\definecolor{appaccent}{RGB}{42,76,108}
\providecommand{\SEL}{\cellcolor{appgreen}}
\definecolor{refgray}{RGB}{110,118,130}
\providecommand{\refrow}[1]{\textcolor{refgray}{#1}}
\ifdefined\corollary\else\newtheorem{corollary}[lemma]{Corollary}\fi
\ifdefined\restatedproposition\else\newtheorem*{restatedproposition}{Proposition~\ref{prop:force}}\fi

\raggedbottom
\makeatletter
\setlength{\@fptop}{0pt}
\setlength{\@fpsep}{18pt plus 4pt}
\setlength{\@fpbot}{0pt plus 1fil}
\makeatother
\renewcommand{\topfraction}{0.9}
\renewcommand{\bottomfraction}{0.8}
\renewcommand{\textfraction}{0.08}
\renewcommand{\floatpagefraction}{0.7}

\addtocontents{toc}{\protect\setcounter{tocdepth}{2}}
\begingroup
\small
\renewcommand{\contentsname}{Appendix Contents}
\hypersetup{colorlinks=true,linkcolor=appaccent}
\pdfbookmark[0]{Appendix Contents}{appendix.contents}
\tableofcontents
\endgroup
\clearpage

\section{Controller Map and Interval Construction}
\label{app:derivation}
\subsection{Impedance Controller}
\label{app:impedance}
This appendix expands the controller map used in Section~\ref{sec:admissible}. The Franka arm uses FORGE's task-space impedance controller~\citep{noseworthy2025forge}, built on Factory~\citep{narang2022factory}. Its seven-dimensional action contains six pose channels and a success prediction. FORGE fixes roll and pitch, so translation and yaw determine physical motion.

The controller smooths each action as $\tilde a_t=\lambda a_t+(1-\lambda)\tilde a_{t-1}$. The translational target is measured relative to the fixed part. A clipped target error and a damping term produce the commanded wrench. Applying the actuation dead zone gives

\begin{align}
    p^{\mathrm{d}}_t &= p^{\mathrm{ref}} + B\, \tilde a_{t,1:3}, \qquad
    W_t = K_p \clip\big(p^{\mathrm{d}}_t - p_t;\, \pm\delta\big) - K_d\, v_t, \label{eq:controller}\\
    W^{\mathrm{app}}_t &= D_{d}(W_t), \qquad D_{d}(w) = \sign(w)\, \max\big(0,\, |w| - d\big), \nonumber
\end{align}
Here $p_t$ and $v_t$ are fingertip position and velocity. The action scale $B$, stiffness $K_p$, damping $K_d$, and position-error clip $\delta$ are diagonal. The target is held for $\Delta t=1/15$\,s. The controller recomputes its wrench and joint torques at 120\,Hz during that period.

Each episode randomizes the gains, clip, smoothing factor $\lambda\sim\mathcal U[0.025,0.1]$, and dead-zone widths $d_i\sim\mathcal U[0,5)$\,N. Dead zones are also redrawn every 2\,s. The agent observes sensor histories. CAP reads the realized parameters through the controller interface.

We collect the quantities needed for the affine wrench map in $x_t=(K_p,K_d,\lambda,d,\delta,\tilde a_{t-1},p_t,v_t,p^{\mathrm{ref}})$. The speed and redraw guards additionally use geometry, inertia, and the redraw schedule, as detailed in Appendix~\ref{app:state}.

\subsection{Admissible Intervals}
\label{app:intervals}
\textbf{Affine wrench model.}
When the target-error clip is inactive, smoothing and impedance control give an affine map on each translational axis. Passing it through the dead zone gives the applied wrench
\begin{equation}
    W_{t,i}(a) = c_i(x_t) + m_i(x_t)\, a_i, \qquad
    W^{\mathrm{app}}_{t,i} = D_{d_i}\big(W_{t,i}(a)\big),
    \label{eq:wrench}
\end{equation}
with $c_i(x_t) = K_{p,i}\big(p^{\mathrm{ref}}_i + B_i (1-\lambda)\, \tilde a_{t-1,i} - p_{t,i}\big) - K_{d,i}\, v_{t,i}$
and $m_i(x_t) = K_{p,i} B_i \lambda > 0$. All quantities in \eqref{eq:wrench} are read from the
controller at the decision instant, and none is learned.

\textbf{Force interval.}
The dead zone can absorb several newtons of commanded wrench. Accounting for that absorption allows useful actuation within an applied-wrench budget. We assign the step budget
$\bar F_t \le F_{\max} - \varepsilon$ for the step and an allocation $\kappa \in \mathbb{R}^3_{>0}$,
$\|\kappa\|_2 \le 1$ ($\kappa_i = 1/\sqrt{3}$ throughout),
\begin{equation}
    \mathcal{I}^{\mathrm{force}}_i(x_t) = \Big\{ a_i : \big| c_i + m_i a_i \big| \le
    d_i + \big(\kappa_i\, \bar F_t - K_{p,i}\, |v_{t,i}|\, \Delta t\big)_+ \Big\},
    \label{eq:force}
\end{equation}
where $\Delta t$ is the control period and $K_{p,i}|v_{t,i}|\Delta t$ reserves a drift allowance under the current-velocity approximation. The within-period extension requires the drift bound stated in Appendix~\ref{app:proof}. By \eqref{eq:wrench}, every $a_i$ in this
interval gives $|W^{\mathrm{app}}_{t,i}| \le \kappa_i \bar F_t$ and hence $\|W^{\mathrm{app}}_t\|_2 \le \bar F_t$.
The budget is tightened by the force measured at the previous step,
$\bar F_t = \min\big(\bar F,\, (\beta\, \bar F + (1-\beta)(F_{\max} - \|F_{t-1}\|))_+\big)$ with
$\bar F = F_{\max} - \varepsilon$ and $\beta \in (0,1]$: the full budget is available while the
previous reading stays within $\varepsilon$, and it decreases linearly as that reading grows. The margin $\varepsilon$
is calibrated to the observed difference between the modeled applied wrench and the sensor reading
per task from traces of separate calibration runs (Appendices~\ref{app:guard} and~\ref{app:calibration}).

\textbf{Speed interval.}
Approach motion can produce an impact before the next decision. CAP therefore uses a local free-body model to limit speed near the mating surface. With directional inertia $\Lambda_i$, its predicted next velocity is
$v_{t,i} + \Delta t\, W^{\mathrm{app}}_{t,i}/\Lambda_i$ after one control period. Requiring its magnitude to be at most $\bar v_{t,i}$ confines the applied wrench to
$[\ell_i, u_i] = \big[-\Lambda_i (\bar v_{t,i} + v_{t,i})/\Delta t,\; \Lambda_i (\bar v_{t,i} - v_{t,i})/\Delta t\big]$
and, through the dead zone and the affine map, the action to
\begin{equation}
    \mathcal{I}^{\mathrm{vel}}_i(x_t) = \Big\{ a_i : \ell_i \le D_{d_i}\big(c_i + m_i a_i\big) \le u_i \Big\},
    \label{eq:vel}
\end{equation}
an interval because $D_d$ is monotone (Lemma~\ref{lem:deadzone}). The budget $\bar v_t$ is a braking
curve of the distance to the surface (Appendix~\ref{app:calibration}): it equals the touchdown speed
$v_{\mathrm{touch}} = F_{\mathrm{imp}}/\sigma$ within one step of the surface, where $\sigma$ is the
calibrated force overshoot per unit approach speed and $F_{\mathrm{imp}}$ the share of the limit reserved
for impact, and grows with the distance as the speed from which the controller can still brake to
$v_{\mathrm{touch}}$.

\textbf{Projection.} The clip in \eqref{eq:controller} makes \eqref{eq:wrench} exact only where it
is inactive; since
$K_{p,i}\,(p^{\mathrm{d}}_{t,i} - p_{t,i}) = c_i + K_{d,i}\, v_{t,i} + m_i a_i$, a third interval,
$\mathcal{I}^{\mathrm{clip}}_i(x_t) = \{ a_i : | c_i + K_{d,i}\, v_{t,i} + m_i a_i | \le K_{p,i}\, \delta_i \}$,
keeps it inactive by construction. Since $m_i > 0$, each of the three constraints is an interval in
the single coordinate $a_i$, so the admissible set is a box and the Euclidean projection onto it is a
componentwise clamp:
\begin{align}
    \mathcal{A}(x_t) &= \prod_{i=1}^{3} \Big( \mathcal{I}^{\mathrm{clip}}_i(x_t) \cap
    \mathcal{I}^{\mathrm{vel}}_i(x_t) \cap \mathcal{I}^{\mathrm{force}}_i(x_t) \cap [-1,1] \Big)
    \times [-1,1]^{\,n-3}, \label{eq:admissible}\\
    a^{\mathrm{exec}}_t &= \Pi_{\mathcal{A}(x_t)}(a_t) = \clip\big(a_t;\, \mathcal{A}(x_t)\big).
    \label{eq:projection}
\end{align}
The translational constraints give a product of intervals. On a nonempty intersection, coordinate-wise clipping is its Euclidean projection. The gear keep-out and nut seating rules additionally constrain yaw (Appendix~\ref{app:keepout}).

The implementation applies the target-clip, speed, and force intervals in sequence, followed by the action range. Lemma~\ref{lem:clamp} proves that these operations equal the projection when the intersection is nonempty. With an empty intersection, their order specifies a priority rule. Corollary~\ref{cor:unsaturated} states the remaining force-bound conditions in that case.

All interval endpoints use closed-form arithmetic. The layer requires an explicit controller map and a constant number of operations per axis. Earlier closed-form corrections include local safety layers~\citep{dalal2018safe} and constrained network outputs~\citep{min2025hardnet}. Here, the axis-wise controller map makes each physical constraint an interval in a single action coordinate.

\FloatBarrier

\section{Proofs}
\label{app:proof}

The proof has three steps. We invert the dead-zone map, establish the projection identity for intersecting intervals, and combine interval membership with the controller law. The final subsection states the additional conditions for measured force and within-period motion.

\subsection{Notation and Assumptions}
\label{app:proof-setting}

Fix a decision instant $t$. All quantities below are read from the same controller state $x_t$,
and we drop the dependence on $x_t$ where it is unambiguous; Table~\ref{tab:notation} lists the
symbols. For each translational axis $i\in\{1,2,3\}$ we assume
\begin{equation}
\begin{gathered}
K_{p,i}>0,\qquad B_i>0,\qquad \lambda>0,\qquad d_i\ge0,\qquad \delta_i\ge0,\qquad
\Lambda_i>0,\qquad \bar v_{t,i}\ge0,\\
\kappa_i>0,\qquad \textstyle\sum_{j=1}^{3}\kappa_j^2\le1,\qquad
0\le\bar F_t\le F_{\max}-\varepsilon .
\end{gathered}
\label{eq:app-assumptions}
\end{equation}
Each condition holds by construction: the action scale is a positive constant, the gains, the
smoothing factor and the clip are drawn from positive ranges (Appendix~\ref{app:tasks}), the speed
budget is at least the touchdown speed (Appendix~\ref{app:calibration}), $\kappa_i=1/\sqrt3$, and
the force budget is clamped to $[0,F_{\max}-\varepsilon]$ (Appendix~\ref{app:guard}). The first
three conditions give $m_i=K_{p,i}B_i\lambda>0$, so the map $a_i\mapsto c_i+m_ia_i$
of~\eqref{eq:wrench} is strictly increasing. $\Lambda_i$ is the inertia used by the speed interval
(Appendix~\ref{app:state}). We write $\eta_i=K_{p,i}|v_{t,i}|\Delta t$ for the drift allowance
of~\eqref{eq:force} and $b_i=d_i+(\kappa_i\bar F_t-\eta_i)_+$ for the half-width of the force
interval, so that $\mathcal I^{\mathrm{force}}_i=\{a_i:|c_i+m_ia_i|\le b_i\}$. All statements
concern the translational wrench at the decision instant and the one-step free-body velocity model
of Section~\ref{sec:admissible}.

\begin{table}[!htbp]
\centering
\caption{\textbf{Notation.} Vectors have three translational components unless stated otherwise.
Controller parameters and state are evaluated at the current decision instant.}
\label{tab:notation}
\small
\begin{tabular*}{\linewidth}{@{}l@{\extracolsep{\fill}}p{0.72\linewidth}@{}}
\toprule
Symbol & Meaning \\
\midrule
$a_t,\ a^{\mathrm{exec}}_t$ & Proposed and executed actions in $[-1,1]^7$ \\
$\tilde a_{t-1},\ \lambda$ & Previous smoothed action and action-smoothing factor \\
$p_t,\ v_t,\ p^{\mathrm{ref}}$ & Fingertip position, fingertip velocity and position reference \\
$K_p,\ K_d,\ B,\ \delta$ & Stiffness, damping, action scale and position-error clip (diagonal) \\
$c_i,\ m_i$ & Intercept and slope of the commanded wrench, $W_{t,i}=c_i+m_ia_i$ \\
$d_i,\ D_{d_i}$ & Dead-zone width and dead-zone map, $W^{\mathrm{app}}_{t,i}=D_{d_i}(W_{t,i})$ \\
$F_t,\ F_{\max},\ \varepsilon$ & Wrist force, force limit and calibrated margin \\
$\bar F_t,\ \kappa_i$ & Force budget of the step and its share on axis $i$ \\
$\eta_i,\ b_i$ & Drift allowance $K_{p,i}|v_{t,i}|\Delta t$ and half-width of the force interval \\
$\Lambda_i,\ \bar v_{t,i}$ & Inertia used by the speed interval and speed budget of axis $i$ \\
$\ell_i,\ u_i$ & Bounds on the applied wrench imposed by the speed budget \\
$L_i,\ U_i$ & Endpoints of the admissible action interval of axis $i$ \\
$\Delta t$ & Control period, $1/15$\,s \\
\bottomrule
\end{tabular*}
\end{table}

\subsection{Auxiliary Lemmas}
\label{app:proof-lemmas}

The first lemma gives exact preimages under the dead zone. The second establishes when sequential clipping equals projection onto the intersection.

\begin{lemma}[Dead zone]
\label{lem:deadzone}
Let $d\ge0$ and $D_d(w)=\sign(w)\,(|w|-d)_+$.
\textup{(i)} $D_d$ is continuous, nondecreasing and $1$-Lipschitz, and $|D_d(w)|=(|w|-d)_+$.
\textup{(ii)} For $\ell\le u$, define
\begin{equation}
g_d^-(\ell)=
\begin{cases}\ell-d,&\ell\le0,\\ \ell+d,&\ell>0,\end{cases}
\qquad
g_d^+(u)=
\begin{cases}u-d,&u<0,\\ u+d,&u\ge0.\end{cases}
\label{eq:app-deadzone-endpoints}
\end{equation}
Then
\begin{equation}
\{w:\ell\le D_d(w)\le u\}=\big[g_d^-(\ell),\,g_d^+(u)\big],
\label{eq:app-deadzone-preimage}
\end{equation}
and this interval is nonempty.
\textup{(iii)} For $m>0$ and $c\in\mathbb R$,
$\{a:\ell\le D_d(c+ma)\le u\}=\big[(g_d^-(\ell)-c)/m,\,(g_d^+(u)-c)/m\big]$.
\end{lemma}

\begin{proof}
On its three branches,
\begin{equation}
D_d(w)=
\begin{cases}
w+d,&w<-d,\\
0,&-d\le w\le d,\\
w-d,&w>d,
\end{cases}
\label{eq:app-deadzone-branches}
\end{equation}
the map is affine with slopes $1$, $0$ and $1$ and is continuous at $w=\pm d$, which gives (i).
For (ii), consider the lower bound first. If $\ell>0$, then $D_d(w)\ge\ell$ holds only on the
positive branch, where it reads $w\ge\ell+d$. If $\ell\le0$, it holds on the zero and positive
branches and, on the negative branch, exactly when $w\ge\ell-d$; since $\ell-d\le-d$, the set is
$\{w\ge\ell-d\}$. In both cases the lower bound is $w\ge g_d^-(\ell)$. The same argument turns
$D_d(w)\le u$ into $w\le u-d$ if $u<0$ and into $w\le u+d$ if $u\ge0$, that is,
$w\le g_d^+(u)$. The interval is nonempty: if $\ell\le0\le u$, then
$g_d^-(\ell)\le-d\le d\le g_d^+(u)$; if $u<0$, then $g_d^-(\ell)=\ell-d\le u-d$; and if $\ell>0$,
then $\ell+d\le u+d$. Part (iii) follows from (ii) because $a\mapsto c+ma$ is strictly increasing
for $m>0$.
\end{proof}

For $\ell=u=0$, \eqref{eq:app-deadzone-preimage} is the whole dead band $[-d,d]$, endpoints
included. The force interval is the case $u_i=-\ell_i=(\kappa_i\bar F_t-\eta_i)_+$ of
Lemma~\ref{lem:deadzone}(iii), for which $g^-_{d_i}(\ell_i)=-b_i$ and $g^+_{d_i}(u_i)=b_i$:
\begin{equation}
\mathcal I^{\mathrm{force}}_i
=\big\{a_i:\ |D_{d_i}(c_i+m_ia_i)|\le(\kappa_i\bar F_t-\eta_i)_+\big\}.
\label{eq:app-force-applied}
\end{equation}
Applying Lemma~\ref{lem:deadzone}(iii) to the three constraints of Section~\ref{sec:admissible}
gives their endpoints in closed form:
\begin{align}
L_i^{\mathrm{force}}&=\frac{-b_i-c_i}{m_i}, &
U_i^{\mathrm{force}}&=\frac{b_i-c_i}{m_i},\label{eq:app-force-endpoints}\\
L_i^{\mathrm{vel}}&=\frac{g_{d_i}^-(\ell_i)-c_i}{m_i}, &
U_i^{\mathrm{vel}}&=\frac{g_{d_i}^+(u_i)-c_i}{m_i},\label{eq:app-velocity-endpoints}\\
L_i^{\mathrm{clip}}&=\frac{-K_{p,i}\delta_i-c_i-K_{d,i}v_{t,i}}{m_i}, &
U_i^{\mathrm{clip}}&=\frac{K_{p,i}\delta_i-c_i-K_{d,i}v_{t,i}}{m_i},
\label{eq:app-clip-endpoints}
\end{align}
with $\ell_i=-\Lambda_i(\bar v_{t,i}+v_{t,i})/\Delta t$ and $u_i=\Lambda_i(\bar v_{t,i}-v_{t,i})/\Delta t$
as in~\eqref{eq:vel}. Since $u_i-\ell_i=2\Lambda_i\bar v_{t,i}/\Delta t\ge0$, the speed interval is
nonempty. When $v_{t,i}>\bar v_{t,i}$, both bounds on the applied wrench are negative and the
interval demands braking; the sign cases of~\eqref{eq:app-deadzone-endpoints} preserve this demand.
The admissible interval of axis $i$ is $[L_i,U_i]$ with
\begin{equation}
L_i=\max\{-1,L_i^{\mathrm{clip}},L_i^{\mathrm{vel}},L_i^{\mathrm{force}}\},\qquad
U_i=\min\{1,U_i^{\mathrm{clip}},U_i^{\mathrm{vel}},U_i^{\mathrm{force}}\},
\label{eq:app-intersection}
\end{equation}
and the translational part of $\mathcal A(x_t)$ is nonempty if and only if $L_i\le U_i$ for
$i=1,2,3$. Equality is allowed and fixes the action of that axis.

\begin{lemma}[Sequential clipping]
\label{lem:clamp}
Let $I_j=[l_j,u_j]$, $j=1,\dots,k$, be closed intervals with nonempty intersection $I=[L,U]$, and
let $C_I(q)=\min\{U,\max\{L,q\}\}$. Then
\begin{equation}
C_{I_k}\circ\cdots\circ C_{I_1}(q)=C_I(q)
\qquad\text{for every }q\in\mathbb R.
\label{eq:app-clamp-composition}
\end{equation}
Consequently, clipping each translational coordinate through the action range, the target-clip,
speed and force intervals, and the action range again yields the Euclidean
projection~\eqref{eq:projection} whenever $\mathcal A(x_t)\ne\emptyset$.
\end{lemma}

\begin{proof}
Let $I_1$ and $I_2$ intersect and write $L=\max\{l_1,l_2\}\le U=\min\{u_1,u_2\}$. If
$q\in[L,U]$, neither clip moves $q$. Let $q<L$. If $L=l_2$, then $C_{I_1}(q)\le C_{I_1}(L)=L$ by
monotonicity and $L\in I_1$, so the second clip returns $L$. If $L=l_1>l_2$, then
$C_{I_1}(q)=L\in I_2$, and the second clip leaves it. The case $q>U$ is symmetric. Hence
$C_{I_2}\circ C_{I_1}=C_{I_1\cap I_2}$. Every partial intersection $I_1\cap\dots\cap I_j$ contains
$I$ and is therefore nonempty, so induction on $k$ gives~\eqref{eq:app-clamp-composition}. For the
projection, $\mathcal A(x_t)$ is a product of intervals and $\|y-a\|_2^2=\sum_i(y_i-a_i)^2$
separates over coordinates; on $[L_i,U_i]$ the unique minimizer of $(y_i-a_i)^2$ is
$C_{[L_i,U_i]}(a_i)$. Repeating the action range does not change the intersection.
\end{proof}

\subsection{Proof of Proposition~\ref{prop:force}}
\label{app:proof-prop}

\begin{restatedproposition}[Controller-output bound]
Let $\mathcal{A}(x_t)\neq\emptyset$. For every proposed action $a_t\in\mathcal{A}$, the executed
action $a^{\mathrm{exec}}_t=\Pi_{\mathcal{A}(x_t)}(a_t)$ leaves the target clip inactive, so that
the commanded wrench is given by~\eqref{eq:wrench}; the applied wrench satisfies
$|W^{\mathrm{app}}_{t,i}|\le\kappa_i\bar F_t$ for $i=1,2,3$ and hence
$\|W^{\mathrm{app}}_t\|_2\le\bar F_t\le F_{\max}-\varepsilon$; and the free-body velocity after the
step satisfies $|v_{t,i}+\Delta t\,W^{\mathrm{app}}_{t,i}/\Lambda_i|\le\bar v_{t,i}$.
\end{restatedproposition}

\begin{proof}
We first verify the affine controller branch. Since $\mathcal A(x_t)\neq\emptyset$, the executed action lies in every interval
of~\eqref{eq:admissible}; for the sequential clip of the implementation this is
Lemma~\ref{lem:clamp}. Let $\tilde a_{t,i}=\lambda a^{\mathrm{exec}}_{t,i}+(1-\lambda)\tilde
a_{t-1,i}$ be the smoothed action. By the definition of $c_i$ and $m_i$ below~\eqref{eq:wrench},
\begin{equation}
c_i+K_{d,i}v_{t,i}+m_ia^{\mathrm{exec}}_{t,i}
=K_{p,i}\big(p^{\mathrm{ref}}_i+B_i\tilde a_{t,i}-p_{t,i}\big)
=K_{p,i}\big(p^{\mathrm d}_{t,i}-p_{t,i}\big).
\label{eq:app-unclipped}
\end{equation}
Membership in $\mathcal I^{\mathrm{clip}}_i$ bounds the left side by $K_{p,i}\delta_i$ in absolute
value, and dividing by $K_{p,i}>0$ gives $|p^{\mathrm d}_{t,i}-p_{t,i}|\le\delta_i$. On this range
the clip in~\eqref{eq:controller} is the identity, endpoints included, so the commanded wrench is
exactly $W_{t,i}=c_i+m_ia^{\mathrm{exec}}_{t,i}$.

Next, membership in $\mathcal I^{\mathrm{force}}_i$ gives $|W_{t,i}|\le b_i$, and
Lemma~\ref{lem:deadzone}(i) gives
\begin{equation}
|W^{\mathrm{app}}_{t,i}|=(|W_{t,i}|-d_i)_+\le(b_i-d_i)_+=(\kappa_i\bar F_t-\eta_i)_+
\le\kappa_i\bar F_t ,
\label{eq:app-force-proof}
\end{equation}
where the last step uses $\eta_i\ge0$ and $\kappa_i\bar F_t\ge0$. Summing squares,
$\|W^{\mathrm{app}}_t\|_2^2\le\bar F_t^2\sum_i\kappa_i^2\le\bar F_t^2$, and
$\bar F_t\le F_{\max}-\varepsilon$ by~\eqref{eq:app-assumptions}.

Finally, membership in $\mathcal I^{\mathrm{vel}}_i$ and Lemma~\ref{lem:deadzone}(iii) give
$\ell_i\le W^{\mathrm{app}}_{t,i}\le u_i$. Multiplying by $\Delta t/\Lambda_i>0$ and adding
$v_{t,i}$,
\begin{equation}
-\bar v_{t,i}\le v_{t,i}+\frac{\Delta t}{\Lambda_i}W^{\mathrm{app}}_{t,i}\le\bar v_{t,i},
\end{equation}
which is the velocity bound.
\end{proof}

The same argument applies when the dead-zone guard of Appendix~\ref{app:guard} replaces $b_i$ by a
smaller half-width $b^{\mathrm{guard}}_i\le b_i$, since $|W_{t,i}|\le b^{\mathrm{guard}}_i$ implies
$|W_{t,i}|\le b_i$. For a zero budget, $\bar F_t=0$, each force interval requires
$|W_{t,i}|\le d_i$ and the modeled applied wrench is zero; the construction never divides by the
budget.

\subsection{Remarks on the Assumptions}
\label{app:proof-practice}

Proposition~\ref{prop:force} assumes a nonempty admissible set and bounds the modeled applied
wrench at the decision instant. We now consider steps whose intervals do not intersect, the
relation to the measured force, the motion within a control period, and the inertia used by the
speed interval.

\textbf{Empty intersections.}
Each interval has ordered endpoints, by $K_{p,i}\delta_i\ge0$ for the target-clip interval,
Lemma~\ref{lem:deadzone}(ii) for the speed interval and $b_i\ge0$ for the force interval, so only
their intersection can be empty. The implementation always returns an action through the ordered
clips of Lemma~\ref{lem:clamp}: target clip, speed, force, and then the action range. When the
intervals of an axis do not intersect, the force interval therefore takes precedence over the speed
and target-clip intervals, and the action range takes precedence over all three.

\begin{corollary}[Force bound on unsaturated steps]
\label{cor:unsaturated}
Let $x_t$ satisfy~\eqref{eq:app-assumptions}, and let $a^{\mathrm{exec}}_t$ be the output of the
ordered clips, whether or not $\mathcal A(x_t)$ is empty. If the final clip to the action range
does not change the translational coordinates, and
$a^{\mathrm{exec}}_{t,i}\in\mathcal I^{\mathrm{clip}}_i(x_t)$ for $i=1,2,3$, then
$|W^{\mathrm{app}}_{t,i}|\le\kappa_i\bar F_t$ for $i=1,2,3$ and
$\|W^{\mathrm{app}}_t\|_2\le\bar F_t\le F_{\max}-\varepsilon$.
\end{corollary}

\begin{proof}
Each force interval is nonempty because $b_i\ge0$, so the output of the force clip lies in
$\mathcal I^{\mathrm{force}}_i$. By hypothesis the final clip leaves it unchanged, so
$a^{\mathrm{exec}}_{t,i}\in\mathcal I^{\mathrm{force}}_i\cap\mathcal I^{\mathrm{clip}}_i$. The first
two parts of the proof of Proposition~\ref{prop:force} use only these two memberships.
\end{proof}

The evaluation logs the two failure conditions as action-range saturation and target-clip infeasibility (Appendix~\ref{app:metrics}). Speed infeasibility alone can leave the force bound intact because force has priority. In the main-table evaluations, target-clip infeasibility is zero. The corollary therefore applies on their unsaturated steps (Table~\ref{tab:diagnostics}).

\textbf{Measured force.}
The limit in~\eqref{eq:problem} applies to the wrist force read at the decision instants. We index
the reading by the step it follows: $F_t$ is read at the end of the period in which
$a^{\mathrm{exec}}_t$ acts, and $F_{t-1}$, the reading available when $a^{\mathrm{exec}}_t$ is
chosen, sets the budget $\bar F_t$. On every step covered by Proposition~\ref{prop:force} or
Corollary~\ref{cor:unsaturated}, a measured force that exceeds the modeled applied wrench by at most
the margin satisfies the limit:
\begin{equation}
\|F_t\|_2\le\|W^{\mathrm{app}}_t\|_2+\varepsilon\le\bar F_t+\varepsilon\le F_{\max}.
\label{eq:app-sensor-condition}
\end{equation}
The first inequality relates the measured wrist-force signal to the modeled controller output. Appendix~\ref{app:calibration} estimates the margin from separate traces of their positive difference. This empirical calibration supports the measured-force evaluation under the reported conditions. The speed constraint additionally reduces approach motion that can generate contact transients. Main-table metrics and Figure~\ref{fig:exceedance} evaluate the filtered wrist force directly.

\textbf{Motion within the control period.}
Proposition~\ref{prop:force} concerns the wrench at the decision instant, while the controller
evaluates~\eqref{eq:controller} at $120$\,Hz against a moving arm with the target held. At the
current velocity the proportional term changes by at most $\eta_i$ over one period, and the force
interval reserves this amount. For the within-period extension, the drift bound applies to the total change in commanded wrench, including proportional and damping terms. If the commanded wrench within the period is $W_{t,i}+e_i$ with
$|e_i|\le\eta_i\le\kappa_i\bar F_t$ and the dead zone does not change, Lemma~\ref{lem:deadzone}(i)
and~\eqref{eq:app-force-proof} give
\begin{equation}
|D_{d_i}(W_{t,i}+e_i)|\le|D_{d_i}(W_{t,i})|+|e_i|\le(\kappa_i\bar F_t-\eta_i)+\eta_i
=\kappa_i\bar F_t .
\label{eq:app-drift-bound}
\end{equation}
A change of the dead zone within the period is handled by the guard of Appendix~\ref{app:guard}.

\textbf{Scaled inertia.}
The speed interval uses $\Lambda_i=\rho\,\Lambda^{\mathrm{arm}}_i$, a scaled copy of the arm's
task-space inertia $\Lambda^{\mathrm{arm}}_i$ with $0<\rho\le1$ (Appendix~\ref{app:state}). Let
$v_i^+=v_i+\Delta t\,W_i^{\mathrm{app}}/\Lambda_i$ be the velocity that the interval bounds. The
free-body update with the arm's inertia is
\begin{equation}
v_i+\frac{\Delta t}{\Lambda^{\mathrm{arm}}_i}W_i^{\mathrm{app}}=(1-\rho)\,v_i+\rho\,v_i^+ .
\label{eq:app-inertia}
\end{equation}
If $v_i$ and $v_i^+$ both lie in $[-\bar v_i,\bar v_i]$, so does their convex combination. The
scaling is therefore conservative for a state within its budget; for a state above it, the speed
interval demands the recovery of the scaled model.

\FloatBarrier

\section{Implementation Details}
\label{app:implementation}

This section describes how the projection is computed and calibrated, the task-specific yaw
constraints, and how the two backbones learn with the projection in the loop.

\subsection{Controller State and Computational Cost}
\label{app:state}

CAP reads the controller interface immediately before execution. It uses the realized gains, action smoothing and filter memory, dead-zone widths, target clips, and fingertip pose and velocity. The speed interval also uses the held part's height above the fixed part and directional inertia. The redraw guard uses the known simulator schedule. Gains, smoothing and filter memory are software controller quantities; pose and velocity are kinematic readings; the speed interval additionally uses task geometry and directional inertia. The feedback budget reads the filtered wrist force before the artificial observation noise. These interface quantities are specified separately from the task-dependent calibration in Appendix~\ref{app:calibration}.

The inertia is obtained from the translational Jacobian $J$ and joint-space mass matrix $M_q$

\begin{equation}
\Lambda^{\mathrm{arm}}_i=\left[(JM_q^{-1}J^\top)_{ii}\right]^{-1},\qquad
\Lambda_i=\rho\,\Lambda^{\mathrm{arm}}_i,\qquad \rho=0.8 .
\label{eq:app-inertia-estimate}
\end{equation}
It is computed from the arm model at every step and is not learned. Algorithm~\ref{alg:cap-collection}
summarizes one control step of \modelname[-WM], including what is written to the replay buffer.
\modelname[-PPO] executes the same projection; it stores the raw Gaussian sample for its likelihood
ratio (Appendix~\ref{app:action-credit}).

\begin{algorithm}[ht]
\caption{One control step of \modelname[-WM] during training}
\label{alg:cap-collection}
\begin{algorithmic}[1]
\Require Observation history $o_t$, controller state $x_t$, wrist force $F_{t-1}$
\State Propose $a_t$: the first action of the MPPI solution of~\eqref{eq:mpc}, with exploration noise
\State Compute $c_i$, $m_i$, $\eta_i$, the force budget~\eqref{eq:app-feedback}, the guard~\eqref{eq:app-scheduled-guard} and the speed budget~\eqref{eq:app-braking-root}
\State Form $\mathcal I^{\mathrm{clip}}_i$, $\mathcal I^{\mathrm{vel}}_i$, $\mathcal I^{\mathrm{force}}_i$ from~\eqref{eq:app-force-endpoints}--\eqref{eq:app-clip-endpoints}
\State $q\gets\clip(a_t;[-1,1]^7)$
\For{$i=1,2,3$}
  \State $q_i\gets C_{\mathcal I^{\mathrm{force}}_i}\big(C_{\mathcal I^{\mathrm{vel}}_i}(C_{\mathcal I^{\mathrm{clip}}_i}(q_i))\big)$
\EndFor
\State $a^{\mathrm{exec}}_t\gets\clip(q;[-1,1]^7)$; apply the task's yaw constraint, if any (Appendix~\ref{app:keepout})
\State Execute $a^{\mathrm{exec}}_t$; observe $r_t$, $o_{t+1}$ and $F_t$
\State Store $(o_t,a_t,r_t,o_{t+1})$ in the replay buffer, with $a^{\mathrm{exec}}_t$ beside it
\State Append $a^{\mathrm{exec}}_t$ and the new controller reading to the observation history
\end{algorithmic}
\end{algorithm}

\textbf{Computational cost.}
The interval arithmetic is batched over environments and uses a constant number of elementwise
operations per axis. On one CPU thread (Intel Core i9-13900KF, PyTorch 2.7), one call of the layer
on a batch of $128$ environments, including its diagnostics, takes $0.34$ to $0.39$\,ms (median of
$3{,}000$ calls), under $0.6\%$ of the $66.7$\,ms control period. The inertia
estimate~\eqref{eq:app-inertia-estimate} adds one $7\times7$ linear solve per environment. Over a
full evaluation of $640$ episodes on the same machine, \modelname[-WM] took $848$\,s against
$858$\,s for TD-MPC2 + depth on peg insertion, and $1{,}126$\,s against $1{,}137$\,s on gear meshing.

\subsection{Force Budget and Dead-Zone Guard}
\label{app:guard}

With $\bar F=F_{\max}-\varepsilon$, the force budget of a step is
\begin{equation}
\bar F_t=\min\big\{\bar F,\ \max\{0,\ \beta\bar F+(1-\beta)(F_{\max}-\|F_{t-1}\|_2)\}\big\},
\qquad \beta=0.5,
\label{eq:app-feedback}
\end{equation}
where $\|F_{t-1}\|_2$ is the wrist force at the end of the previous period, read before
observation noise. The first decision of an episode uses $\bar F$. If $\|F_{t-1}\|_2\le\varepsilon$,
the full budget is available; above that, the budget decreases linearly in the measured force and
reaches zero at $\|F_{t-1}\|_2=F_{\max}+\bar F$. In every case $0\le\bar F_t\le\bar F$,
as~\eqref{eq:app-assumptions} requires.

FORGE redraws the dead-zone widths of all environments every $2$\,s, that is, every $30$ decisions.
The dead-zone credit $d_i$ in $b_i$ is valid only for the current width: a commanded wrench that a
wide dead zone absorbs is transmitted almost in full once the width is redrawn small. We therefore
bound the commanded wrench ahead of each redraw. Let $j_t\ge0$ be the number of decisions until the
last decision before the redraw ($j_t=0$ at that decision). With rate $\zeta$ and slack $s$,
\begin{equation}
S_{t,i}=\kappa_i(\bar F-s)_++j_t\,\zeta\,m_i,
\qquad
b_i^{\mathrm{guard}}=\min\big\{b_i,\ (S_{t,i}-\eta_i)_+\big\},
\label{eq:app-scheduled-guard}
\end{equation}
and $b_i^{\mathrm{guard}}$ replaces $b_i$ in~\eqref{eq:app-force-endpoints}. All reported
configurations use $\zeta=1$ and $s=0$. Since $b_i^{\mathrm{guard}}\le b_i$, the guard only
tightens the force interval, and Proposition~\ref{prop:force} and Corollary~\ref{cor:unsaturated}
hold unchanged. At $j_t=0$, $|W_{t,i}|\le\kappa_i\bar F$ whatever the current width, and since
$|D_{d'}(w)|\le|w|$ for every $d'\ge0$, the same wrench satisfies the per-axis bound after any
redraw. The bound $S_{t,i}$ grows by $\zeta m_i$ per decision of lead time, so the guard binds only
in a short window before each redraw.

\subsection{Calibration and the Speed Budget}
\label{app:calibration}

Calibration sets four quantities. The margin $\varepsilon$ and impact slope $\sigma$ are task dependent. Gear meshing additionally uses cap- and backbone-specific values. The gap margin $g_0$ and deceleration ceiling $a_{\max}$ are fixed per task.

All estimates use traced 640-episode evaluations of separate calibration runs. Those runs use CAP-PPO with projection disabled or an earlier CAP controller on the same task. Table~\ref{tab:calibration} gives the resulting values. The following paragraphs specify each estimator.

\textbf{Margin.}
A trace row $t$ records the modeled applied-wrench norm $w_t=\|D_d(c+m\,a^{\mathrm{exec}}_t)\|_2$,
the fingertip speed $s_t$ and the height $g_t$, all read at decision $t$, and the wrist-force norm
$f_t$ read at the end of the same period. Let $e_t=(f_t-w_t)_+$. Slow-contact rows satisfy
$f_t\ge0.35$\,N and $s_t\le0.02$\,m/s; free-space rows satisfy $f_t<0.35$\,N, $g_t>5$\,mm and
$w_t<0.05$\,N. The margin is the $99$th percentile of $e_t$ over slow-contact rows plus the $99$th
percentile of $f_t$ over free-space rows plus $0.005$\,N; for peg insertion,
$0.646+0.133+0.005=0.784$\,N.

\textbf{Impact slope and touchdown speed.}
A contact onset is a row at which $f_t$ crosses $0.35$\,N from below. Its peak is the larger force of
the onset row and the following row, and its approach speed is the speed at the preceding decision.
For onsets approached at $5$\,mm/s or faster, we divide the positive excess of the peak over the
modeled applied wrench by the approach speed; $\sigma$ is the $99$th percentile of this ratio. The
slope therefore describes the part of the force peak that the applied-wrench model does not account
for. The touchdown speed is $v_{\mathrm{touch}}=F_{\mathrm{imp}}/\sigma$, where the impact allowance
$F_{\mathrm{imp}}$ is fixed at about $47\%$ of $F_{\max}-\varepsilon$ in every configuration.

\textbf{Gap margin and deceleration ceiling.}
The gap margin $g_0$ is the $99$th percentile of the height reading at the first contact onset of
each episode, clipped at zero and rounded up to the next $0.5$\,mm; it places the touchdown speed at
the heights where first contact occurs. The braking curve assumes a constant deceleration, whereas
the action filter limits how fast the commanded wrench can change. We therefore cap the deceleration
at $a_{\max}$, estimated from pre-contact rows in which the speed exceeds its budget: the mean
deceleration over such rows of an episode, at the $25$th percentile over episodes, rounded down to a
multiple of $0.005$\,m/s$^2$. This gives $a_{\max}=0.02$\,m/s$^2$ on peg insertion; on gear meshing
and nut threading we use the estimator's floor of $0.005$\,m/s$^2$. The braking capacity of the
controller itself, $(K_{p,i}\delta_i-d_i)/(2\Lambda_i)$, exceeds $a_{\max}$ in every reported
configuration, so $a_{\max}$ sets the curve.

\begin{figure}[!htbp]
\centering
\includegraphics[width=\linewidth]{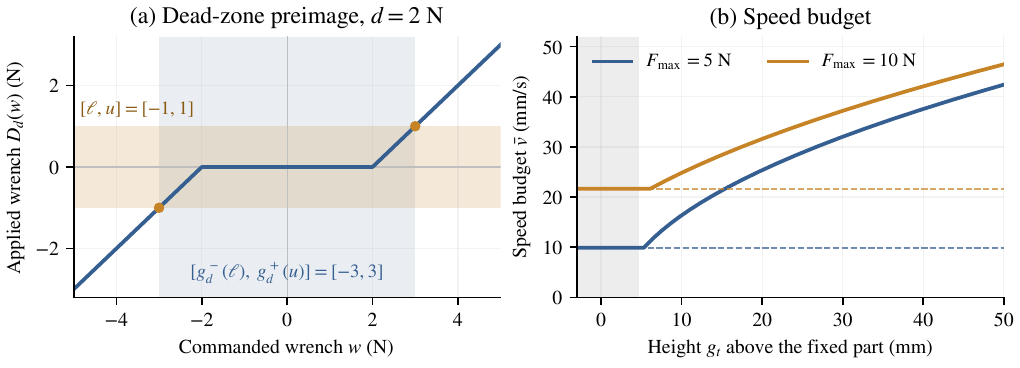}
\caption{\textbf{Interval construction.} (a)~Preimage of the applied-wrench interval $[-1,1]$\,N
under a dead zone of $d=2$\,N (Lemma~\ref{lem:deadzone}); the commanded-wrench interval contains the
whole zero-output band. (b)~Speed budget~\eqref{eq:app-braking-root} for peg insertion at $5$ and
$10$\,N with $\|v_t\|_2=0.01$\,m/s; dashed lines mark the touchdown speeds, and the shaded region
has zero effective gap. Both panels evaluate the formulas of this appendix.}
\label{fig:appendix-construction}
\end{figure}

\textbf{Speed budget.}
With the effective gap $g_t^{\mathrm{eff}}=(g_t-g_0-\|v_t\|_2\Delta t)_+$, the speed budget is
\begin{equation}
\begin{aligned}
\bar v_{t}&=\min\big\{v_{\mathrm{ceil}},\ \max\{v_{\mathrm{touch}},v^*_{t}\}\big\},\\
v^*_{t}&=-a_{\max}\Delta t+\sqrt{(a_{\max}\Delta t)^2+v_{\mathrm{touch}}^2+2a_{\max}g_t^{\mathrm{eff}}}\,,
\end{aligned}
\label{eq:app-braking-root}
\end{equation}
with $v_{\mathrm{ceil}}=0.25$\,m/s; the budget is the same on the three axes. The root $v^*_t$ is the
nonnegative solution of $v^2=v_{\mathrm{touch}}^2+2a_{\max}(g_t^{\mathrm{eff}}-v\Delta t)$: a part
approaching at the budget speed can still brake to $v_{\mathrm{touch}}$ after one further period of
travel, and $g_t^{\mathrm{eff}}$ already discounts the distance $\|v_t\|_2\Delta t$ that the part
covers before the next decision. Since $v^*_t\ge v_{\mathrm{touch}}$ exactly when
$g_t^{\mathrm{eff}}\ge v_{\mathrm{touch}}\Delta t$, the budget equals $v_{\mathrm{touch}}$ near the
surface and grows with the gap beyond it (Figure~\ref{fig:appendix-construction}(b)). The budget enters
the interval through~\eqref{eq:app-velocity-endpoints}. For gear meshing at $5$\,N,
\modelname[-WM] uses $v_{\mathrm{touch}}=14.6$\,mm/s below $g_t=-6$\,mm, where the held gear is
already engaged, and $F_{\mathrm{imp}}/\sigma$ above; every other configuration uses a single
touchdown speed.

\begin{table}[!htbp]
\centering
\caption{\textbf{Projection parameters.} Calibrated margin $\varepsilon$, enforced bound
$F_{\max}-\varepsilon$ on the modeled applied wrench, impact allowance $F_{\mathrm{imp}}$, impact
slope $\sigma$, gap margin $g_0$ and deceleration ceiling $a_{\max}$
(Appendix~\ref{app:calibration}). Where the two backbones differ, the entry reads
\modelname[-PPO]\,/\,\modelname[-WM].}
\label{tab:calibration}
\footnotesize
\begin{tabular*}{\linewidth}{@{}l@{\extracolsep{\fill}}ccccccc@{}}
\toprule
& $F_{\max}$ & $\varepsilon$ & $F_{\max}-\varepsilon$ & $F_{\mathrm{imp}}$ & $\sigma$ & $g_0$ & $a_{\max}$ \\
\textbf{Task} & (N) & (N) & (N) & (N) & (N$\cdot$s/m) & (mm) & (m/s$^2$) \\
\midrule
\multirow{2}{*}{Peg insertion} & 5 & 0.784 & 4.216 & 2.000 & 201.7 & 4.0 & 0.02 \\
 & 10 & 0.784 & 9.216 & 4.372 & 201.7 & 4.0 & 0.02 \\
\midrule
\multirow{2}{*}{Gear meshing} & 5 & 0.893\,/\,0.899 & 4.107\,/\,4.101 & 1.948 & 211.1\,/\,266.3 & 27.5 & 0.005 \\
 & 10 & 1.138 & 8.862 & 4.204 & 266.3 & 27.5 & 0.005 \\
\midrule
\multirow{2}{*}{Nut threading} & 5 & 0.680 & 4.320 & 2.049 & 251.1 & 27.0 & 0.005 \\
 & 10 & 0.680 & 9.320 & 4.421 & 251.1 & 27.0 & 0.005 \\
\bottomrule
\end{tabular*}
\end{table}

\subsection{Task-Specific Yaw Constraints}
\label{app:keepout}

The yaw command is the sixth action coordinate. Its smoothed target is
\begin{equation}
\psi^{\mathrm d}=O-S\big[\lambda a_{t,6}+(1-\lambda)\tilde a_{t-1,6}\big],
\qquad O=\pi/4,\quad S=3\pi/4 ,
\label{eq:app-yaw-map}
\end{equation}
which is strictly decreasing in $a_{t,6}$. A target interval $[\psi_-,\psi_+]$ therefore has the
action preimage
\begin{equation}
\left[
\frac{(O-\psi_+)/S-(1-\lambda)\tilde a_{t-1,6}}{\lambda},\
\frac{(O-\psi_-)/S-(1-\lambda)\tilde a_{t-1,6}}{\lambda}
\right].
\label{eq:app-yaw-preimage}
\end{equation}
The yaw constraints clip this coordinate to the preimage and then to $[-1,1]$. They act on a
coordinate that the translational intervals do not involve, so Proposition~\ref{prop:force} and its
proof are unchanged.

\textbf{Gear meshing.}
The gripper holds the gear by its hub beside two free gears. When the fingers descend below neighboring hub tops, some yaw angles bring a finger into contact with a hub. The resulting rotational interaction can create force peaks outside the translational model. The keep-out rule restricts the target yaw to avoid those angles.

Let $\psi_h$ be hand yaw, $\psi_r$ the gear-row heading, and $\psi_f=\pi/2$ the finger-line direction in the hand frame. The wrapped relative angle is $\vartheta=\operatorname{wrap}_{[-\pi/2,\pi/2)}(\psi_h+\psi_f-\psi_r)$. The admissible target offsets are

\begin{equation}
[\psi_--\psi_h,\ \psi_+-\psi_h]=
\begin{cases}
[\alpha_{\min}-\vartheta,\ \pi-\alpha_{\min}-\vartheta],&\vartheta\ge0,\\
[-\pi+\alpha_{\min}-\vartheta,\ -\alpha_{\min}-\vartheta],&\vartheta<0.
\end{cases}
\label{eq:app-yaw-arc}
\end{equation}
Both arcs have width $\pi-2\alpha_{\min}$ and keep the finger line at least $\alpha_{\min}$ from the
row; inside the cone, the nearer edge demands motion out of it. Both gear \modelname[-WM]
configurations use $\alpha_{\min}=42^\circ$: the largest finger-line angle at which we measured a
hub-contact torque, $37.0^\circ$, plus a closed-loop overshoot allowance of $4.8^\circ$, rounded up.
The keep-out is active from the first decision of an episode and constrains the target yaw, which the
rotational controller tracks.

\textbf{Nut threading.}
Nut threading requires both thread engagement and a final hand yaw below zero. The available yaw range permits a finite amount of screwing-in rotation. Using that rotation before the nut reaches the bolt can reduce later progress.

For CAP-WM at 5\,N, the seat-first gate restricts screwing-in rotation while the nut is above the bolt tip. When $g_t>0$, it requires $\psi^{\mathrm d}\ge\psi_h$. Substituting the yaw map gives

\begin{equation}
a_{t,6}\le
\frac{(O-\psi_h)/S-(1-\lambda)\tilde a_{t-1,6}}{\lambda},
\label{eq:app-seat-gate}
\end{equation}
and the gate releases the coordinate at $g_t\le0$. The other
configurations use no yaw constraint.

\subsection{Learning with the Projection in the Loop}
\label{app:action-credit}

Let $S_x(a)$ denote the complete execution map, including the fallback of
Appendix~\ref{app:proof-practice} and any yaw constraint. If the state $s$ includes the controller
memory, an agent that proposes $a$ collects experience from the system
\begin{equation}
P_S(s'\mid s,a)=P\big(s'\mid s,S_{x(s)}(a)\big),
\qquad R_S(s,a)=R\big(s,S_{x(s)}(a)\big).
\label{eq:app-composed-system}
\end{equation}
Under partial observability the encoder represents the observation history;
\eqref{eq:app-composed-system} specifies which action variable conditions the transition,
independently of the quality of that representation.

\textbf{World-model replay.}
\modelname[-WM] stores the proposal $a_t$ as the action of a transition and records the executed
action beside it. The dynamics, reward and value heads and the policy prior are therefore trained on
the proposal, the variable over which MPPI searches. The observation history receives the executed
command instead, since that command drove the controller. The same convention holds in the seed
phase and in the planning phase, and every environment action passes through the projection. The
planner searches over proposal sequences in the learned model, and the projection is applied to the
selected first action using the current controller state.

\textbf{Policy likelihoods.}
For \modelname[-PPO], let $a_t^{\mathrm{raw}}$ be the Gaussian sample before the clip to the action
range and the projection. The importance ratio is
\begin{equation}
r_t(\theta)=
\frac{\pi_\theta(a_t^{\mathrm{raw}}\mid h_t)}
     {\pi_{\theta_{\mathrm{old}}}(a_t^{\mathrm{raw}}\mid h_t)},
\label{eq:app-ppo-ratio}
\end{equation}
where $h_t$ is the policy's observation history. For a fixed execution map, the trajectory density
factors into policy densities of the raw samples and transition kernels of the
form~\eqref{eq:app-composed-system}, and only the policy factors depend on $\theta$. The
logarithmic derivative therefore gives the usual score terms
$\sum_t\nabla_\theta\log\pi_\theta(a_t^{\mathrm{raw}}\mid h_t)$, which justifies evaluating the ratio
at the recorded raw sample while computing advantages from the rewards of the projected rollout. It
also avoids assigning a Gaussian density to the executed action, whose distribution has mass on the
interval boundaries.

\FloatBarrier

\section{Experimental Setup}
\label{app:config}

This section describes the tasks, observations, rewards, baselines, backbone hyperparameters, training and checkpoint selection, evaluation protocols, metrics, and compute.

\subsection{Tasks and Controller}
\label{app:tasks}

We use the peg-insertion, gear-meshing and nut-threading tasks of
FORGE~\citep{noseworthy2025forge} in Isaac Lab, with FORGE's task-space impedance controller,
rewards and success criteria. Episodes last $10$, $20$ and $30$\,s. The controller runs at
$120$\,Hz and holds each command for eight physics steps, a $15$\,Hz decision rate. The
seven-dimensional action contains translation, rotation and a success prediction; FORGE fixes roll
and pitch, so translation and yaw are the physical control channels.

At every reset, the translational stiffness (nominally $565$\,N/m), the position-error clip
(nominally $0.02$\,m) and the rotation-error clip (nominally $0.097$\,rad) are multiplied or
divided, independently per axis, by a factor drawn uniformly from $[1,1.41]$, $[1,1.25]$ and
$[1,1.29]$, respectively. The smoothing factor is drawn from $\mathcal U[0.025,0.1]$. Translational
dead-zone widths are drawn from $\mathcal U[0,5)$\,N and rotational ones from
$\mathcal U[0,1)$\,N$\cdot$m, at reset and again every $2$\,s. The projection reads the realized
values.

\begin{figure}[!htbp]
\centering
\includegraphics[width=\linewidth]{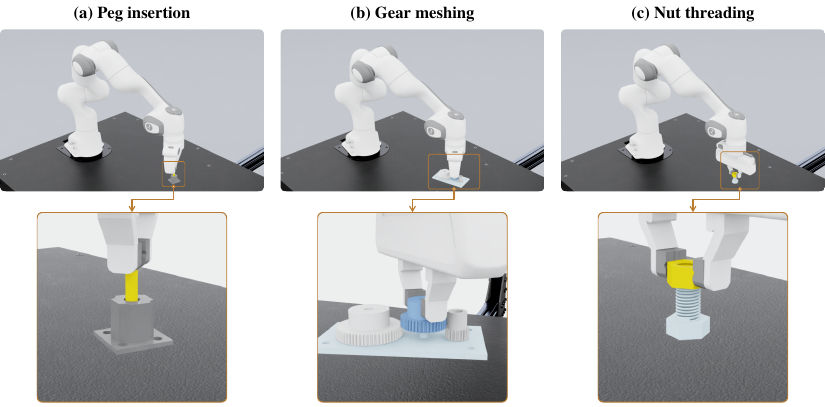}
\caption{\textbf{Simulated assembly scenes and contact geometry.} Top panels show the complete Franka Panda and workpiece. Bottom panels show external close-up views of the mating geometry at the same state. Columns illustrate (a) peg insertion, (b) gear meshing, and (c) nut threading. Frames come from the first CAP-WM episode, seed 0, at 4.0, 6.4, and 26.6\,s respectively. Presentation lighting and a neutral floor material are applied after reaching the selected state. Images are cropped for display. The policy observes wrist depth.}
\label{fig:simulation-scenes}
\end{figure}

\FloatBarrier

\subsection{Observations and Wrist-Force Measurement}
\label{app:observations}
Table~\ref{tab:environment-settings} summarizes the task-specific and shared settings.

The $24$-dimensional observation contains the fingertip position relative to the fixed part, the
fingertip orientation quaternion, the linear and angular velocity, the wrist force, the
force-penalty threshold of the FORGE reward (drawn per episode from $[5,10]$\,N) and the previous
smoothed action. The observed force carries Gaussian noise with a standard deviation of $1$\,N per
axis. The force budget and all metrics use the environment's low-pass filtered wrist force before
this noise.

\textbf{Wrist-force signal and sampling.}
Let $f_k^{\mathrm{wrist}}$ be the translational wrist-force vector from the simulator's incoming-joint-wrench interface at physics update $k$. FORGE updates its filtered wrench in the world frame using
\[
\widetilde f_k^{\mathrm{wrist}}
=\alpha_F f_k^{\mathrm{wrist}}+(1-\alpha_F)\widetilde f_{k-1}^{\mathrm{wrist}},
\qquad \alpha_F=0.25.
\]
The translational vector is expressed in the task frame and read at the end of each eight-physics-step control period to obtain $F_t$. The reported norms and excess-force integrals use this signal before the artificial observation noise is added. The physics-rate channel refers to the same wrist-sensing interface before low-pass filtering. This force filter is distinct from the action-smoothing factor $\lambda$ used in the controller map.

\textbf{Wrist depth.}
The world-model agents receive a $96\times96$ depth image from a wrist-mounted camera with a
$48$\,mm focal length. Metric depth is normalized as
\begin{equation}
I^{\mathrm{norm}}=\clip\left(\frac{I-0.35\,\mathrm m}{0.25\,\mathrm m};\,-1,1\right).
\label{eq:app-depth-normalization}
\end{equation}
The depth encoder has four convolutions with $32$ channels, kernels $7,5,3,3$, stride $2$ and ReLU
activations; its $512$ output features are mapped linearly to the latent dimension and normalized
with SimNorm. A random shift with padding $4$ is applied in gradient updates only. The state and
depth embeddings have equal size and are summed and passed through a fusion MLP,
following~\citet{hansen2026insertion}. The vector input stacks the observation and the executed
action over the last $3$ decisions on peg insertion and the last $32$ on gear meshing and nut
threading ($93$ and $992$ dimensions), together with the current depth frame.

\begin{table}[!htbp]
\centering\footnotesize
\caption{\textbf{Environment and observation settings.} Task-specific settings are listed in the order used throughout the paper. Shared sensing and control settings apply to all three tasks.}
\label{tab:environment-settings}
\begin{tabular*}{\linewidth}{@{}l@{\extracolsep{\fill}}ccc@{}}
\toprule
\textbf{Setting} & \textbf{Peg insertion} & \textbf{Gear meshing} & \textbf{Nut threading}\\
\midrule
Episode duration (s) & 10 & 20 & 30\\
Observation history (decisions) & 3 & 32 & 32\\
Stacked vector input dimension & 93 & 992 & 992\\
Contact-penalty coefficient & 0.2 & 0.05 & 0.05\\
\midrule
\multicolumn{4}{@{}l}{\textbf{Shared control and sensing}}\\
Decision / physics rate & \multicolumn{3}{c}{15 / 120\,Hz (eight physics steps per decision)}\\
Physical action channels & \multicolumn{3}{c}{Translation and yaw; roll and pitch fixed}\\
Observation / action dimension & \multicolumn{3}{c}{24 / 7}\\
Wrist depth / focal length & \multicolumn{3}{c}{$96\times96$ / 48\,mm}\\
Force-filter coefficient $\alpha_F$ & \multicolumn{3}{c}{0.25 (updated at the physics rate)}\\
Force observation noise & \multicolumn{3}{c}{Gaussian, 1\,N standard deviation per axis}\\
Reward force threshold & \multicolumn{3}{c}{Sampled from $[5,10]$\,N during training}\\
Evaluation force caps & \multicolumn{3}{c}{5 and 10\,N, before observation noise}\\
\bottomrule
\end{tabular*}
\end{table}

\FloatBarrier

\subsection{Rewards and Supervision}
\label{app:rewards}
All agents retain the FORGE task reward. Factory supplies geometric alignment and task-progress terms. FORGE adds an asset-relative action penalty, a penalty on filtered force above a sampled threshold, and success-prediction error. The action coefficient is 0.001. The contact-penalty coefficient is 0.2 for peg insertion and 0.05 for gear meshing and nut threading.
The reward is
\begin{equation}
r_t=r_t^{\mathrm{Factory}}-c_a\,e_t^{\mathrm{asset}}
-c_F\big(\|F_t\|_2-F_{\mathrm{thr}}\big)_+-c_{\mathrm{pred},t}\,e_t^{\mathrm{pred}},
\label{eq:app-reward}
\end{equation}
where $r_t^{\mathrm{Factory}}$ collects the Factory terms, $e_t^{\mathrm{asset}}$ is the position and yaw
offset of the commanded target from the fingertip, each normalized by its action threshold, $c_a$ and $c_F$
are the action and contact-penalty coefficients, $F_{\mathrm{thr}}$ is the sampled force threshold, and
$e_t^{\mathrm{pred}}$ is the absolute success-prediction error with activation
$c_{\mathrm{pred},t}\in\{0,1\}$.

The seventh action coordinate predicts success after rescaling to $[0,1]$. Its reward term activates once the configured success fraction reaches 0.25. Success uses the original task conditions, including engagement and rotation for nut threading. Training samples the reward's force threshold from $[5,10]$\,N. Evaluation scores each policy against the fixed cap in the table.

The world model is trained on collected transitions with latent-consistency, reward-prediction, and temporal-difference targets. Wrist depth supplies geometric observations during those interactions.

\FloatBarrier

\subsection{Baselines}
\label{app:baselines}

\textbf{Unconstrained learners.}
PPO (FORGE) follows Isaac Lab's rl\_games recipe. Both actor and critic use a two-layer LSTM with 1,024 units followed by an ELU MLP of widths 512, 128, and 64. Training uses discount 0.995, GAE 0.95, learning rate $10^{-4}$ with adaptive KL threshold 0.008, PPO clip 0.2, four mini-epochs, minibatch size 512, and 128 environments. Peg, gear, and nut use 233, 467, and 351 epochs, with rollout horizons 128, 128, and 256.

TD-MPC2 + depth uses CAP-WM's world-model hyperparameters, observations, and interaction budget. Its execution omits CAP, and its training omits the auxiliary force-prediction heads. Both unconstrained learners are evaluated once per task and scored at both force limits.

\textbf{Constrained reinforcement learning.}
CPO, PID-Lagrangian and FOCOPS~\citep{achiam2017cpo,stooke2020pid,zhang2020focops} maximize the
task reward subject to a budget $d$ on the expected episode cost, with the per-step cost
\begin{equation}
c_t=(\|F_t\|_2-F_{\max})_+\,\Delta t ,
\label{eq:app-cost}
\end{equation}
whose undiscounted episode sum is the excess force of~\eqref{eq:app-evaluation}. Each member follows
its reference implementation with its own networks and optimizer (Table~\ref{tab:saferl-hyper}),
uses $128$ environments for $200$ iterations of one episode per environment, and keeps the FORGE
reward. The budget of a task and force limit is $0.18$ times the excess force of the PPO (FORGE)
policy of Table~\ref{tab:main} at that limit, which gives $d=0.4363$ and $0.02201$\,N$\cdot$s on peg
insertion, $2.332$ and $0.4093$ on gear meshing, and $6.643$ and $0.9836$ on nut threading, at $5$ and
$10$\,N. All three members constrain the undiscounted episode cost. For each
member we evaluate the checkpoint ranked best on training rollouts (by cost feasibility, then
violation, success and return) and the final one. CPO evaluations are available for all three tasks and both force limits.

\begin{table}[!htbp]
\centering
\caption{\textbf{Constrained-RL hyperparameters.} All members share the cost~\eqref{eq:app-cost},
$128$ environments and $200$ iterations. $J_C$: episode cost; $\lambda$, $\nu$: Lagrange
multipliers.}
\label{tab:saferl-hyper}
\footnotesize
\setlength{\tabcolsep}{4pt}
\newcolumntype{Y}{>{\raggedright\arraybackslash}p{0.235\linewidth}}
\begin{tabular*}{\linewidth}{@{}l@{\extracolsep{\fill}}YYY@{}}
\toprule
& \textbf{CPO} & \textbf{PID-Lagrangian} & \textbf{FOCOPS} \\
\midrule
Policy network & LSTM $2\times1{,}024$, MLP $512,128,64$ & MLP $512,512$, LSTM $512$ & MLP $64,64$ \\
Value networks & MLP $512,128,64$ & shared with the policy & MLP $64,64$ \\
Discount (reward / cost) & $0.995$ / $1$ & $0.99$ / $0.99$ & $0.99$ / $1$ \\
GAE (reward / cost) & $0.95$ / $1$ & $0.97$ / $0.97$ & $0.95$ / $0.95$ \\
Constraint on $J_C$ & undiscounted & undiscounted & undiscounted \\
Policy update & KL step $0.01$, $10$ CG iterations & clip $0.1$, $8$ epochs & KL bound $0.02$, $10$ epochs \\
Learning rate & line search ($0.8$, $15$ steps) & $10^{-4}$ & $3\times10^{-4}$, linear decay \\
Multiplier & --- & $K_P=0$, $K_I=1$, $K_D=0$, $\lambda_0=0$ & $\nu_0=0$, rate $0.01$, $\nu\le2$ \\
Other & damping $10^{-5}$ & cost scale $10$ & temperature $1.5$ \\
\bottomrule
\end{tabular*}
\end{table}

\textbf{Safety layers.}
SC$^3$~\citep{wang2025guarding} and FCSCC~\citep{liang2023fcscc} apply execution-time filtering to the fixed PPO (FORGE)
policy of each task in Table~\ref{tab:main}, using a force threshold of $F_{\max}$ and zero additional margin. SC$^3$ solves a quadratic program at
every control step that projects the nominal action onto the half-space of a first-order control
barrier function on the force, with a disturbance observer; it acts in contact. FCSCC projects the
commanded wrench in closed form onto the half-space of a second-order exponential barrier on
$F_{\max}-\|F_t\|_2$, using the task-space inertia read from the simulator, and acts in and out of
contact. Both use a contact stiffness identified on the task. We evaluate several stiffness settings on every task and limit (SC$^3$: the identified
value scaled by $0.25$, $1$ and $4$; FCSCC: the identified value, a stiffer prior and a rigid prior
of $25.4$\,kN/m), with further settings on peg insertion, and report the best setting at each limit
(Table~\ref{tab:families}).

\FloatBarrier

\subsection{Backbone Hyperparameters}
\label{app:hyper}

Table~\ref{tab:network-hyper} lists the hyperparameters of both backbones; the controller
randomization is given in Appendix~\ref{app:tasks}. \modelname[-WM] keeps the training objective,
the model sizes and the planner of TD-MPC2~\citep{hansen2024tdmpc2} in the configuration of its 5M
model and adds the depth encoder and fusion of Appendix~\ref{app:observations}; it also keeps TD-MPC2's
heuristic discount and the burst of updates at the end of the seed phase, during which
uniformly random proposals are executed through the projection. \modelname[-WM] also trains small
force-prediction heads on the detached latent with a separate optimizer; they do not affect the world
model, the planner or the projection. \modelname[-PPO] uses the recurrent actor of the FORGE PPO
recipe, trained from random initialization with the projection active from the first iteration and
the ratio~\eqref{eq:app-ppo-ratio}.

\begin{table}[!htbp]
\centering
\caption{\textbf{Backbone hyperparameters.} Where a value differs by task, the entry reads
peg\,/\,gear\,/\,nut. A vector step advances all $32$ environments of a world-model run by one
decision.}
\label{tab:network-hyper}
\footnotesize
\begin{tabular*}{\linewidth}{@{}l@{\extracolsep{\fill}}l@{}}
\toprule
\multicolumn{2}{@{}l}{\textbf{\modelname[-WM]}}\\
\midrule
Latent / hidden / state-encoder width & $512$ / $512$ / $256$ \\
State-encoder layers / SimNorm group size & $2$ / $8$ \\
Depth encoder & $4$ convolutions ($32$ channels, kernels $7,5,3,3$, stride $2$) \\
Q-functions / Q-function dropout & $5$ / $0.01$ \\
Reward and value bins / range & $101$ / $[-10,10]$ in symlog space \\
Discount & $0.967$ / $0.983$ / $0.989$ (TD-MPC2 heuristic) \\
Learning rate / encoder multiplier & $3\times10^{-4}$ / $0.3$ \\
Consistency / reward / value loss weights & $20$ / $0.1$ / $0.1$ \\
Temporal loss discount $\rho$ / target update rate & $0.5$ / $0.01$ \\
Gradient-norm clip / policy entropy coefficient & $20$ / $10^{-4}$ \\
Planning horizon / iterations / samples & $3$ / $6$ / $512$ \\
Elites / policy-prior trajectories / temperature & $64$ / $24$ / $0.5$ \\
Planning standard deviation (minimum, maximum) & $(0.05,\,2)$ \\
Batch size / gradient updates per vector step & $256$ / $8$ \\
Observation history (decisions) & $3$ / $32$ / $32$ \\
Seed phase (vector steps) & $1{,}000$ / $1{,}500$ / $2{,}250$ \\
Replay capacity (transitions) & $10^6$ / $10^6$ / $5\times10^5$ \\
State input & running normalization, clipped to $\pm10$ \\
Parallel environments & $32$ \\
\midrule
\multicolumn{2}{@{}l}{\textbf{\modelname[-PPO]}}\\
\midrule
Actor & $2$-layer LSTM ($1{,}024$ units), then MLP $512,128,64$ \\
Action distribution & Gaussian, state-dependent standard deviation \\
Discount / GAE parameter & $0.995$ / $0.95$ \\
PPO clip / epochs per iteration & $0.2$ / $4$ \\
Actor / critic learning rate & $10^{-4}$ / $10^{-3}$ \\
Actor learning-rate schedule / KL threshold & adaptive / $0.008$ \\
Actor learning-rate bounds & $[10^{-6},10^{-2}]$ \\
Entropy coefficient / gradient-norm clip & $0$ / $1$ \\
Iterations / environments / episodes per environment & $200$ / $128$ / $1$ \\
\bottomrule
\end{tabular*}
\end{table}

\FloatBarrier

\subsection{Training Budgets and Checkpoint Selection}
\label{app:selection}

Table~\ref{tab:protocol} summarizes the training protocol. The model-free rows use $128$ parallel
environments and the world-model rows $32$, so at equal episode lengths the world-model rows
receive one quarter of the model-free environment steps.

\begin{table}[!htbp]
\centering
\caption{\textbf{Training protocol} of the rows of Table~\ref{tab:main}. Environment steps are in
millions and count every environment of a run. Checkpoint selection is specified below.}
\label{tab:protocol}
\footnotesize
\setlength{\tabcolsep}{3pt}
\begin{tabular*}{\linewidth}{@{}l@{\extracolsep{\fill}}llllll@{}}
\toprule
& \textbf{\begin{tabular}[b]{@{}l@{}}PPO\\(FORGE)\end{tabular}} & \textbf{Safe RL}
& \textbf{\begin{tabular}[b]{@{}l@{}}Safety\\layer\end{tabular}}
& \textbf{\begin{tabular}[b]{@{}l@{}}TD-MPC2\\+ depth\end{tabular}}
& \textbf{\modelname[-PPO]} & \textbf{\modelname[-WM]} \\
\midrule
Learner & PPO & constrained PG & none & TD-MPC2 & PPO & TD-MPC2 \\
Force constraint & none & cost budget & filter & none & projection & projection \\
Observation & proprio. & proprio. & proprio. & + depth & proprio. & + depth \\
Parallel envs.\ & 128 & 128 & --- & 32 & 128 & 32 \\
Env.\ steps, peg & 3.82 & 3.81 & --- & 0.96 & 3.81 & 0.96 \\
Env.\ steps, gear & 7.65 & 7.65 & --- & 1.92 & 7.65 & 1.92 \\
Env.\ steps, nut & 11.50 & 11.49 & --- & 2.88 & 11.49 & 1.21\,/\,2.88 \\

\bottomrule
\end{tabular*}
\end{table}

\textbf{Checkpoint selection.}
For Table~\ref{tab:main}, unconstrained checkpoints are ranked by SR, constrained baselines by SSR with SR as the tie-breaker, and CAP by SR subject to zero recorded violations. The selected CAP candidates also maximize SR over their evaluated candidates.

Candidates include checkpoints, baseline members, and filter settings. Gear and nut world-model comparisons use separate validation episodes; the other main-table cells use the scoring episodes for selection. Training-seed tests use draw 100 with the checkpoint protocol in Appendix~\ref{app:new-seeds}.

\FloatBarrier

\subsection{Validation and Matched Evaluation Protocol}
\label{app:new-protocol}
Table~\ref{tab:evaluation-protocols} summarizes the evaluation settings. The additional studies use three training seeds wherever stated. Model-free training retains 128 environments and world-model training retains 32. Frozen-policy ablations keep the trained weights fixed and change one execution setting. The no-speed retraining study changes the speed-interval flag and keeps the learning recipe. Checkpoint candidates are fixed before validation on draw 90 and testing on draw 100; the nut CAP-PPO study uses epoch-100 checkpoints as specified in Appendix~\ref{app:new-seeds}. The ablation evaluations rerun Full CAP on the same GPU, driver, and rendering backend as the modified arms. This controls evaluation-platform differences. Training can use different machines; each ablation comparison shares its evaluation platform.

\begin{table}[!htbp]
\centering\footnotesize\setlength{\tabcolsep}{4pt}
\caption{\textbf{Evaluation protocols.} Selection and evaluation settings for each study. Counts are per cell or training seed.}
\label{tab:evaluation-protocols}
\begin{tabular*}{\linewidth}{@{}p{0.19\linewidth}@{\extracolsep{\fill}}p{0.25\linewidth}p{0.34\linewidth}r@{}}
\toprule
\textbf{Study} & \textbf{Models} & \textbf{Selection and evaluation} & \textbf{Episodes}\\
\midrule
Main comparison & Selected models & Checkpoint rules in Appendix~\ref{app:selection} & 640\\
Training-seed tests & Three training seeds & Test draw 100; checkpoints in Appendix~\ref{app:new-seeds} & 640\\
Component ablations & Three seeds; matched evaluation platform & Validation draw 90; test draw 100 & 640\\
Exploratory uncertainty & One retrained seed & Validation draw 191 & 640\\
\bottomrule
\end{tabular*}
\end{table}

\FloatBarrier

\subsection{Evaluation Metrics}
\label{app:metrics}

Each cell is evaluated on $N=640$ episodes, five in each of $128$ parallel environments. Let
$s_e\in\{0,1\}$ be the environment's success criterion at the final step of episode $e$, and let
$f_{e,t}=\|F_{e,t}\|_2$ be the wrist force read at the end of control step $t$. Then
\begin{align}
p_e&=\max_t f_{e,t}, &
v_e&=\mathbf 1\{p_e>F_{\max}\},\nonumber\\
\mathrm{SR}&=\frac1N\sum_e s_e, &
\mathrm{SSR}&=\frac1N\sum_e s_e(1-v_e),\nonumber\\
\mathrm{EF}&=\frac1N\sum_e\sum_t(f_{e,t}-F_{\max})_+\Delta t, &
\mathrm{PF}&=\frac1N\sum_e p_e.
\label{eq:app-evaluation}
\end{align}
The number of violating episodes is $\sum_e v_e$. A fixed-policy evaluation on $640$ independent episodes has binomial standard error at most $1/(2\sqrt{640})=1.98$ percentage points. Table~\ref{tab:main} reports descriptive evaluation counts under the selection protocol in Appendix~\ref{app:selection}. Table~\ref{tab:new-seeds} reports test performance and variability across independently trained policies.

\textbf{Projection statistics.}
For each diagnostic flag, we compute the fraction of control steps flagged within an episode and
average it over episodes. The intervention flag records a change of the translational action by the
projection; the yaw constraints are counted separately. The saturation flag marks steps at which the
final clip to the action range changed the output of the force clip, and the speed and target-clip
infeasibility flags mark steps at which the executed action lies outside the corresponding interval
on at least one axis.

\FloatBarrier

\subsection{Compute}
\label{app:compute}

Every training run used a single NVIDIA RTX~4090 or RTX~A6000 GPU and took between $2$ and $20$
hours, including physics simulation, depth rendering and periodic evaluation; the reported cells used
about $300$ GPU-hours in total. The software is Python~3.11, PyTorch~2.7.0 with CUDA~12.8, Isaac
Sim~5.1.0 and Isaac Lab~2.3.2.

\FloatBarrier

\section{Additional Results}
\label{app:additional}

The results below expand the comparisons in Section~\ref{sec:experiments}. Tables~\ref{tab:families} and~\ref{tab:diagnostics} expand the main comparison; Tables~\ref{tab:new-ablations} and~\ref{tab:new-seeds} report the three-seed studies. The exploratory validation study is reported separately in Appendix~\ref{app:new-audits}.

\par\noindent\begin{minipage}{\linewidth}
\setlength{\parskip}{0.5pc}
\subsection{Baseline Families}
\label{app:families}

Table~\ref{tab:families} lists every Safe RL member and safety layer behind the family rows of
Table~\ref{tab:main}. Among the Safe RL members FOCOPS is the best on every task and at both
limits; among the safety layers, FCSCC at $5$\,N and SC$^3$ at $10$\,N on peg insertion and gear meshing, and
FCSCC on nut threading. The family rows of Table~\ref{tab:main} therefore report the strongest
available member for each condition. For FCSCC, stiffening the prior lowers the number of violating
episodes but also the success rate: on nut threading at $10$\,N, it goes from
$424$ violating episodes and $98.0\%$ success at the identified stiffness to $71$ and $77.5\%$
with the stiff prior and $16$ and $8.6\%$ with the rigid one.

\begin{table}[H]
\centering
\caption{\textbf{Baseline families.} For each Safe RL member, the better of its training-ranked and
final checkpoints; for each safety layer, its best stiffness setting (Appendix~\ref{app:baselines});
both by SSR at each limit, ties broken by SR. Shaded: the entry reported in Table~\ref{tab:main}.
}
\label{tab:families}
\footnotesize
\setlength{\tabcolsep}{7.2pt}
\begin{tabular}{@{}ll ccc ccc}
\toprule
& & \multicolumn{3}{c}{$F_{\max}=5$\,N} & \multicolumn{3}{c}{$F_{\max}=10$\,N} \\
\cmidrule(lr){3-5}\cmidrule(l){6-8}
& & \textbf{SR}\,\up & \textbf{SSR}\,\up & \textbf{Viol.}\,\dn & \textbf{SR}\,\up & \textbf{SSR}\,\up & \textbf{Viol.}\,\dn \\
\textbf{Task} & \textbf{Member} & (\%) & (\%) & (/640) & (\%) & (\%) & (/640) \\
\midrule
\multirow{5}{*}{Peg insertion} & CPO & 0.3 & 0.2 & 499 & 2.3 & 1.9 & 188 \\
 & PID-Lagrangian & 0.0 & 0.0 & 0 & 74.4 & 64.2 & 70 \\
 & FOCOPS & \SEL35.9 & \SEL25.2 & \SEL163 & \SEL77.8 & \SEL77.2 & \SEL5 \\
 & SC$^3$ & 75.8 & 2.7 & 620 & \SEL91.9 & \SEL58.3 & \SEL237 \\
 & FCSCC & \SEL90.2 & \SEL6.1 & \SEL594 & 97.0 & 53.6 & 289 \\
\midrule
\multirow{5}{*}{Gear meshing} & CPO & 20.8 & 2.2 & 587 & 50.8 & 40.6 & 131 \\
 & PID-Lagrangian & 0.0 & 0.0 & 89 & 65.0 & 25.0 & 293 \\
 & FOCOPS & \SEL67.2 & \SEL8.0 & \SEL563 & \SEL58.3 & \SEL50.8 & \SEL85 \\
 & SC$^3$ & 67.2 & 1.2 & 631 & \SEL87.3 & \SEL52.7 & \SEL283 \\
 & FCSCC & \SEL21.7 & \SEL2.2 & \SEL307 & 44.5 & 35.2 & 94 \\
\midrule
\multirow{5}{*}{Nut threading} & CPO & 54.7 & 0.2 & 638 & 71.7 & 44.8 & 249 \\
 & PID-Lagrangian & 0.2 & 0.0 & 6 & 56.2 & 2.7 & 578 \\
 & FOCOPS & \SEL89.5 & \SEL0.3 & \SEL638 & \SEL90.6 & \SEL68.6 & \SEL158 \\
 & SC$^3$ & 86.9 & 0.2 & 639 & 94.2 & 35.3 & 409 \\
 & FCSCC & \SEL45.0 & \SEL1.1 & \SEL602 & \SEL77.5 & \SEL72.0 & \SEL71 \\
\bottomrule
\end{tabular}
\end{table}
\end{minipage}\par\medskip

\par\noindent\begin{minipage}{\linewidth}
\setlength{\parskip}{0.5pc}
\subsection{Distribution of Episode Peak Forces}
\label{app:force-distribution}

For the episode peaks $p_e$ of~\eqref{eq:app-evaluation}, define the empirical exceedance
\begin{equation}
\widehat S(F)=\frac1N\sum_{e=1}^N\mathbf 1\{p_e>F\}.
\label{eq:app-exceedance}
\end{equation}
Its value at $F=F_{\max}$ is the fraction of violating episodes. Figure~\ref{fig:exceedance} shows
$\widehat S$ for every row of Table~\ref{tab:main}. The \modelname{} curves reach zero below the limit
on all three tasks: the largest episode peak over the six \modelname{} evaluations at each limit is
$4.78$\,N at $5$\,N and $8.26$\,N at $10$\,N. The other curves show how the episodes of each baseline
distribute around the limit and reveal the spread of forces behind the aggregate SSR values.

\begin{figure}[H]
\centering
\includegraphics[width=\linewidth]{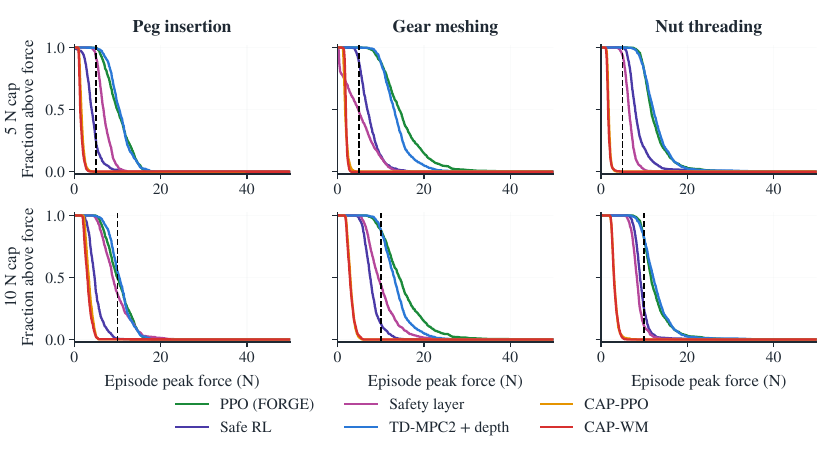}
\caption{\textbf{Episode peak-force exceedance.} Fraction of the $640$ evaluation episodes whose
peak force exceeds the value on the horizontal axis~\eqref{eq:app-exceedance}. Columns are tasks and
rows the two force limits; the height of a curve at the dashed line is the fraction of violating
episodes.}
\label{fig:exceedance}
\end{figure}
\end{minipage}\par\medskip

\par\noindent\begin{minipage}{\linewidth}
\setlength{\parskip}{0.5pc}
\subsection{Component Ablations}
\label{app:component-ablations}
\textbf{Projection during learning.}
The evaluation-only arm applies the complete Full-CAP execution configuration to an unprojected TD-MPC2 checkpoint. This includes gear yaw keep-out and the nut seat gate where configured. Full CAP uses the corresponding learner trained with that layer active.

\textbf{Controller information and dead-zone credit.}
The nominal-parameter arm replaces the randomized controller parameters by nominal values. It retains robot state, geometry, inertia, and the redraw clock. The no-credit arm removes dead-zone credit from the force interval and retains the remaining mechanisms.

Table~\ref{tab:new-ablations} gives the three-seed results. A separate one-seed peg study obtains 98.3\% SSR for Full CAP, 81.1\% for evaluation-only CAP, and 83.0\% for the CAP-WM recipe trained without projection and evaluated with it.

\begin{table}[H]
\centering\footnotesize\setlength{\tabcolsep}{4pt}
\caption{\textbf{Three-seed component ablations at 5\,N.} SR and SSR are mean $\pm$ sample standard deviation in percent. Violations are pooled across 1,920 episodes per row. Green rows mark Full-CAP references; repeated references share the same evaluations.}
\label{tab:new-ablations}
\begin{tabular}{@{}>{\raggedright\arraybackslash}p{\dimexpr0.08\linewidth-1\tabcolsep\relax}>{\raggedright\arraybackslash}p{\dimexpr0.26\linewidth-2\tabcolsep\relax}>{\raggedleft\arraybackslash}p{\dimexpr0.23\linewidth-2\tabcolsep\relax}>{\raggedleft\arraybackslash}p{\dimexpr0.23\linewidth-2\tabcolsep\relax}>{\raggedleft\arraybackslash}p{\dimexpr0.2\linewidth-1\tabcolsep\relax}@{}}
\toprule
\textbf{Task} & \textbf{Arm} & \textbf{SR (\%)} & \textbf{SSR (\%)} & \textbf{Violations}\\
\midrule
\rowcolor{appblue}
\multicolumn{5}{@{}l}{\textbf{Projection during learning}}\\
\addlinespace[2pt]
Gear & Unprojected & 95.6 $\pm$ 0.6 & 0.0 $\pm$ 0.0 & 1919/1920 \\
Gear & CAP at eval & 55.3 $\pm$ 16.2 & 55.3 $\pm$ 16.2 & 0/1920 \\
\rowcolor{appgreen}
Gear & Full CAP & 93.1 $\pm$ 0.8 & 93.1 $\pm$ 0.8 & 0/1920 \\
Nut & Unprojected & 94.1 $\pm$ 3.4 & 0.0 $\pm$ 0.0 & 1920/1920 \\
Nut & CAP at eval & 47.6 $\pm$ 7.6 & 47.6 $\pm$ 7.6 & 0/1920 \\
\rowcolor{appgreen}
Nut & Full CAP & 89.9 $\pm$ 3.0 & 89.9 $\pm$ 3.0 & 0/1920 \\
\midrule
\rowcolor{appblue}
\multicolumn{5}{@{}l}{\textbf{Approach-speed interval}}\\
\addlinespace[2pt]
\rowcolor{appgreen}
Peg & Full CAP & 97.6 $\pm$ 1.0 & 97.6 $\pm$ 1.0 & 0/1920 \\
Peg & Remove at eval & 98.8 $\pm$ 0.3 & 59.1 $\pm$ 2.8 & 770/1920 \\
Peg & Retrained without & 99.2 $\pm$ 0.6 & 56.0 $\pm$ 4.5 & 842/1920 \\
\rowcolor{appgreen}
Gear & Full CAP & 93.1 $\pm$ 0.8 & 93.1 $\pm$ 0.8 & 0/1920 \\
Gear & Remove at eval & 87.5 $\pm$ 2.9 & 35.9 $\pm$ 1.3 & 1133/1920 \\
Gear & Retrained without & 90.1 $\pm$ 0.9 & 31.4 $\pm$ 1.5 & 1254/1920 \\
\midrule
\rowcolor{appblue}
\multicolumn{5}{@{}l}{\textbf{Live controller parameters}}\\
\addlinespace[2pt]
\rowcolor{appgreen}
Peg & Live & 97.6 $\pm$ 1.0 & 97.6 $\pm$ 1.0 & 0/1920 \\
Peg & Nominal & 38.2 $\pm$ 2.3 & 32.4 $\pm$ 2.0 & 155/1920 \\
\rowcolor{appgreen}
Gear & Live & 93.1 $\pm$ 0.8 & 93.1 $\pm$ 0.8 & 0/1920 \\
Gear & Nominal & 28.2 $\pm$ 2.9 & 15.2 $\pm$ 2.6 & 418/1920 \\
\midrule
\rowcolor{appblue}
\multicolumn{5}{@{}l}{\textbf{Dead-zone credit}}\\
\addlinespace[2pt]
\rowcolor{appgreen}
Peg & Credit & 97.6 $\pm$ 1.0 & 97.6 $\pm$ 1.0 & 0/1920 \\
Peg & No credit & 11.1 $\pm$ 0.5 & 11.1 $\pm$ 0.5 & 0/1920 \\
\rowcolor{appgreen}
Nut & Credit & 89.9 $\pm$ 3.0 & 89.9 $\pm$ 3.0 & 0/1920 \\
Nut & No credit & 24.2 $\pm$ 1.7 & 24.2 $\pm$ 1.7 & 0/1920 \\
\bottomrule
\end{tabular}\end{table}
\end{minipage}\par\medskip

\par\noindent\begin{minipage}{\linewidth}
\setlength{\parskip}{0.5pc}
\subsection{Projection Activity}
\label{app:projection-activity}

Table~\ref{tab:diagnostics} reports the projection statistics of every \modelname{} cell. The
projection changes the translational action on $94.6$ to $99.4\%$ of control steps; a proposal
outside a nonempty box is moved to its nearest admissible point, so frequent intervention does not
indicate infeasibility. Target-clip feasibility holds throughout these evaluations. Together with the action-range
clip diagnostics, Corollary~\ref{cor:unsaturated} certifies the modeled-wrench bound on
$93.3$--$98.4\%$ of steps, expressed as episode-averaged fractions. The remaining $1.6$--$6.7\%$
activate the final action-range clip. Measured-force compliance is evaluated over all steps,
including these fallback decisions. The yaw constraints change the yaw
command on $59.7\%$ (gear, $5$\,N), $16.0\%$ (gear, $10$\,N) and $14.9\%$ (nut, $5$\,N) of steps.

\begin{table}[H]
\centering
\caption{\textbf{Projection statistics} on the $640$ evaluation episodes of Table~\ref{tab:main}, as
within-episode step fractions averaged over episodes (Appendix~\ref{app:metrics}). Interv.: the
translational action is changed. Yaw: the gear keep-out or the nut seat gate changes the yaw
command. Sat.: the final action-range clip decides. Speed inf.\ and Clip inf.: the executed action
lies outside the speed or target-clip interval on at least one axis. The flags are not mutually
exclusive.}
\label{tab:diagnostics}
\footnotesize
\begin{tabular*}{\linewidth}{@{}ll@{\extracolsep{\fill}}cccccc@{}}
\toprule
& & \multicolumn{2}{c}{\textbf{Activity}} & \multicolumn{3}{c}{\textbf{Feasibility}} & \\
\cmidrule(lr){3-4}\cmidrule(lr){5-7}
& & Interv. & Yaw & Sat. & Speed inf. & Clip inf. & Viol. \\
\textbf{Task} & \textbf{Method} & (\%) & (\%) & (\%) & (\%) & (\%) & (/640) \\
\midrule
\multirow{4}{*}{Peg insertion} & \modelname[-PPO], 5\,N & 98.9 & --- & 6.73 & 5.27 & 0.00 & 0 \\
 & \modelname[-PPO], 10\,N & 97.6 & --- & 2.92 & 2.85 & 0.00 & 0 \\
 & \modelname[-WM], 5\,N & 97.8 & --- & 6.17 & 5.26 & 0.00 & 0 \\
 & \modelname[-WM], 10\,N & 94.6 & --- & 2.39 & 2.36 & 0.00 & 0 \\
\midrule
\multirow{4}{*}{Gear meshing} & \modelname[-PPO], 5\,N & 98.2 & --- & 2.41 & 2.00 & 0.00 & 0 \\
 & \modelname[-PPO], 10\,N & 98.4 & --- & 1.59 & 1.59 & 0.00 & 0 \\
 & \modelname[-WM], 5\,N & 96.5 & 59.7 & 2.98 & 2.59 & 0.00 & 0 \\
 & \modelname[-WM], 10\,N & 98.3 & 16.0 & 2.12 & 2.12 & 0.00 & 0 \\
\midrule
\multirow{4}{*}{Nut threading} & \modelname[-PPO], 5\,N & 99.0 & --- & 3.65 & 3.61 & 0.00 & 0 \\
 & \modelname[-PPO], 10\,N & 99.4 & --- & 2.02 & 2.02 & 0.00 & 0 \\
 & \modelname[-WM], 5\,N & 98.3 & 14.9 & 3.41 & 3.36 & 0.00 & 0 \\
 & \modelname[-WM], 10\,N & 97.7 & --- & 2.22 & 2.23 & 0.00 & 0 \\
\bottomrule
\end{tabular*}
\end{table}
\end{minipage}\par\medskip

\par\noindent\begin{minipage}{\linewidth}
\setlength{\parskip}{0.5pc}
\subsection{Execution Ablations with Fixed Checkpoints}
\label{app:execution-ablations}

Table~\ref{tab:ablations-full} changes how trained controllers are executed while keeping their
weights. Removing the projection from a trained \modelname{} controller lowers safe success to at
most $0.9\%$ at both limits, and task success falls as well, on every task and for both backbones:
the learned behavior relies on the projection that shaped it during training.

The second comparison places the unconstrained agents behind an evaluation-time force gate with the
margin of the corresponding \modelname{} configuration. While the wrist force read at the last
decision is at or below $F_{\max}-\varepsilon$, the gate passes the proposed translation unchanged;
above it, the gate replaces the translation by $\clip(-c/m;[-1,1])$, the command for which the
controller model predicts zero commanded wrench, and hands control back at the next decision at which
the force is again below the threshold. At $5$\,N the gate reduces the excess force of PPO (FORGE)
from $2.424$ to $0.693$\,N$\cdot$s on peg insertion, from $12.957$ to $1.497$ on gear meshing and
from $36.907$ to $1.565$ on nut threading, yet at least $632$ of $640$ episodes still violate the
limit. The gate reacts to contact that has already occurred, whereas \modelname{} bounds the next
command from the controller state and the approach speed before contact.

\begin{table}[H]
\centering
\caption{\textbf{Fixed-checkpoint execution ablations.} Each modified row reuses the checkpoint of
the grey row above it, which repeats Table~\ref{tab:main}. The execution marked ``none'' removes projection and yaw constraints. The force-gate rows use the evaluation-time gate described above. All weights are fixed. Bold counts indicate zero violating episodes.}
\label{tab:ablations-full}
\footnotesize
\setlength{\tabcolsep}{3pt}
\begin{tabular*}{\linewidth}{@{}c@{\extracolsep{\fill}\hspace{\tabcolsep}}ll*{4}{c}>{\color{black!30}\vrule width 0.4pt}c*{4}{c}@{}}
\toprule
& & & \multicolumn{4}{c}{Force limit $F_{\max}=5$\,N} & \multicolumn{1}{c}{} & \multicolumn{4}{c}{Force limit $F_{\max}=10$\,N} \\
\cmidrule(lr){4-7}\cmidrule(l){9-12}
& & & \textbf{SR}\,\up & \textbf{SSR}\,\up & \textbf{EF}\,\dn & \textbf{Viol.}\,\dn & & \textbf{SR}\,\up & \textbf{SSR}\,\up & \textbf{EF}\,\dn & \textbf{Viol.}\,\dn \\
& \textbf{Policy} & \textbf{Execution} & (\%) & (\%) & (N$\cdot$s) & (/640) & & (\%) & (\%) & (N$\cdot$s) & (/640) \\
\midrule
\multirow{8}{*}{\rotatebox[origin=c]{90}{\textbf{Peg insertion}}} & \refrow{\modelname[-PPO]} & \refrow{projection} & \refrow{78.0} & \refrow{78.0} & \refrow{0.000} & \refrow{\textbf{0}} & & \refrow{90.5} & \refrow{90.5} & \refrow{0.000} & \refrow{\textbf{0}} \\
 & \modelname[-PPO] & none & 32.8 & 0.0 & 53.290 & 640 & & 25.2 & 0.9 & 11.406 & 607 \\
 & \refrow{\modelname[-WM]} & \refrow{projection} & \refrow{98.1} & \refrow{98.1} & \refrow{0.000} & \refrow{\textbf{0}} & & \refrow{98.0} & \refrow{98.0} & \refrow{0.000} & \refrow{\textbf{0}} \\
 & \modelname[-WM] & none & 92.0 & 0.0 & 59.947 & 639 & & 78.8 & 0.6 & 22.118 & 634 \\
 & \refrow{PPO (FORGE)} & \refrow{none} & \refrow{97.7} & \refrow{0.3} & \refrow{2.424} & \refrow{638} & & \refrow{97.7} & \refrow{49.5} & \refrow{0.122} & \refrow{319} \\
 & PPO (FORGE) & force gate & 94.1 & 1.2 & 0.693 & 632 & & 97.0 & 48.3 & 0.123 & 322 \\
 & \refrow{TD-MPC2 + depth} & \refrow{none} & \refrow{99.1} & \refrow{0.6} & \refrow{7.784} & \refrow{636} & & \refrow{99.1} & \refrow{44.7} & \refrow{0.284} & \refrow{352} \\
 & TD-MPC2 + depth & force gate & 94.1 & 0.9 & 1.069 & 634 & & 98.9 & 49.1 & 0.108 & 323 \\
\midrule
\multirow{8}{*}{\rotatebox[origin=c]{90}{\textbf{Gear meshing}}} & \refrow{\modelname[-PPO]} & \refrow{projection} & \refrow{94.7} & \refrow{94.7} & \refrow{0.000} & \refrow{\textbf{0}} & & \refrow{98.4} & \refrow{98.4} & \refrow{0.000} & \refrow{\textbf{0}} \\
 & \modelname[-PPO] & none & 30.0 & 0.0 & 124.159 & 640 & & 17.0 & 0.0 & 59.574 & 640 \\
 & \refrow{\modelname[-WM]} & \refrow{projection} & \refrow{94.8} & \refrow{94.8} & \refrow{0.000} & \refrow{\textbf{0}} & & \refrow{95.0} & \refrow{95.0} & \refrow{0.000} & \refrow{\textbf{0}} \\
 & \modelname[-WM] & none & 34.8 & 0.0 & 141.262 & 640 & & 50.8 & 0.0 & 48.326 & 639 \\
 & \refrow{PPO (FORGE)} & \refrow{none} & \refrow{99.8} & \refrow{0.0} & \refrow{12.957} & \refrow{640} & & \refrow{99.8} & \refrow{11.6} & \refrow{2.274} & \refrow{566} \\
 & PPO (FORGE) & force gate & 88.9 & 0.0 & 1.497 & 640 & & 97.5 & 17.8 & 0.338 & 525 \\
 & \refrow{TD-MPC2 + depth} & \refrow{none} & \refrow{95.2} & \refrow{0.0} & \refrow{20.253} & \refrow{640} & & \refrow{95.2} & \refrow{11.6} & \refrow{1.849} & \refrow{565} \\
 & TD-MPC2 + depth & force gate & 88.3 & 0.0 & 1.533 & 640 & & 94.5 & 21.4 & 0.248 & 501 \\
\midrule
\multirow{8}{*}{\rotatebox[origin=c]{90}{\textbf{Nut threading}}} & \refrow{\modelname[-PPO]} & \refrow{projection} & \refrow{61.4} & \refrow{61.4} & \refrow{0.000} & \refrow{\textbf{0}} & & \refrow{87.2} & \refrow{87.2} & \refrow{0.000} & \refrow{\textbf{0}} \\
 & \modelname[-PPO] & none & 38.0 & 0.0 & 108.685 & 640 & & 27.2 & 0.0 & 74.597 & 639 \\
 & \refrow{\modelname[-WM]} & \refrow{projection} & \refrow{89.2} & \refrow{89.2} & \refrow{0.000} & \refrow{\textbf{0}} & & \refrow{91.9} & \refrow{91.9} & \refrow{0.000} & \refrow{\textbf{0}} \\
 & \modelname[-WM] & none & 58.9 & 0.0 & 205.018 & 640 & & 65.8 & 0.0 & 69.085 & 640 \\
 & \refrow{PPO (FORGE)} & \refrow{none} & \refrow{97.5} & \refrow{0.0} & \refrow{36.907} & \refrow{640} & & \refrow{97.5} & \refrow{18.6} & \refrow{5.465} & \refrow{520} \\
 & PPO (FORGE) & force gate & 88.0 & 0.0 & 1.565 & 640 & & 98.0 & 28.6 & 0.216 & 455 \\
 & \refrow{TD-MPC2 + depth} & \refrow{none} & \refrow{96.2} & \refrow{0.0} & \refrow{43.143} & \refrow{640} & & \refrow{96.2} & \refrow{17.7} & \refrow{3.907} & \refrow{525} \\
 & TD-MPC2 + depth & force gate & 78.1 & 0.2 & 1.379 & 638 & & 87.5 & 13.9 & 0.286 & 535 \\
\bottomrule
\end{tabular*}
\end{table}
\end{minipage}\par\medskip

\par\noindent\begin{minipage}{\linewidth}
\setlength{\parskip}{0.5pc}
\subsection{Wrist Depth and the Projection}
\label{app:depth-study}

Table~\ref{tab:modality} separates the two components of \modelname[-WM] on peg insertion at
$5$\,N: four world models on the same camera-equipped environment, with and without wrist depth and
with and without the projection, all with an observation history of $3$ and eight gradient updates
per vector step. With the projection, depth raises success from $85.2\%$ to $98.1\%$, and neither
controller has a violating episode. Without the projection, depth raises success from $89.7\%$ to
$99.1\%$, while safe success stays at $0.6\%$. Depth improves task completion although the
proprioceptive models were trained on $3.3$ times as many environment steps, and the projection
provides force compliance with or without depth.

\begin{table}[H]
\centering
\caption{\textbf{Wrist depth and the projection} on peg insertion at $F_{\max}=5$\,N. Four TD-MPC2
world models on the same camera-equipped environment; the last two rows are the TD-MPC2 + depth and
\modelname[-WM] entries of Table~\ref{tab:main}. Bold entries mark the highest SSR and lowest EF.}
\label{tab:modality}
\footnotesize
\begin{tabular*}{\linewidth}{@{}cc@{\extracolsep{\fill}}ccccccc@{}}
\toprule
\textbf{Depth} & \textbf{Projection} & Env.\ steps & \textbf{SR}\,\up & \textbf{SSR}\,\up & \textbf{EF}\,\dn & \textbf{PF}\,\dn & \textbf{Viol.}\,\dn \\
& & ($\times 10^6$) & (\%) & (\%) & (N$\cdot$s) & (N) & (/640) \\
\midrule
--- & --- & 3.20 & 89.7 & 0.6 & 6.530 & 10.53 & 636 \\
--- & \yes & 3.20 & 85.2 & 85.2 & \textbf{0.000} & 1.71 & 0 \\
\yes & --- & 0.96 & 99.1 & 0.6 & 7.784 & 10.60 & 636 \\
\yes & \yes & 0.96 & 98.1 & \textbf{98.1} & \textbf{0.000} & 1.56 & 0 \\
\bottomrule
\end{tabular*}
\end{table}
\end{minipage}\par\medskip

\par\noindent\begin{minipage}{\linewidth}
\setlength{\parskip}{0.5pc}
\subsection{Planning Horizon at Inference}
\label{app:horizon}

Table~\ref{tab:horizon} evaluates the $5$\,N \modelname[-WM] checkpoints of Table~\ref{tab:main},
trained with the planning horizon $H=3$ of Table~\ref{tab:network-hyper}, with $H=8$ and $H=15$ at
inference. Success depends on the horizon, and the training horizon is the best or within
$1.4$ points of it on every task. Across the $3{,}840$ additional episodes, the projection
maintains force compliance with zero force-cap violations at either inference horizon.

\begin{table}[H]
\centering
\caption{\textbf{Planning horizon at inference.} The $5$\,N \modelname[-WM] checkpoints of
Table~\ref{tab:main}, trained with $H=3$, evaluated with the horizon of each column on $640$
episodes. The $H=3$ column repeats Table~\ref{tab:main}. Bold counts indicate zero violating episodes.}
\label{tab:horizon}
\footnotesize
\begin{tabular*}{\linewidth}{@{}l@{\extracolsep{\fill}}ccccccccc@{}}
\toprule
& \multicolumn{3}{c}{$H=3$ (training)} & \multicolumn{3}{c}{$H=8$} & \multicolumn{3}{c}{$H=15$} \\
\cmidrule(lr){2-4}\cmidrule(lr){5-7}\cmidrule(l){8-10}
& \textbf{SR}\,\up & \textbf{SSR}\,\up & \textbf{Viol.}\,\dn & \textbf{SR}\,\up & \textbf{SSR}\,\up
& \textbf{Viol.}\,\dn & \textbf{SR}\,\up & \textbf{SSR}\,\up & \textbf{Viol.}\,\dn \\
\textbf{Task} & (\%) & (\%) & (/640) & (\%) & (\%) & (/640) & (\%) & (\%) & (/640) \\
\midrule
Peg insertion & 98.1 & 98.1 & \textbf{0} & 94.7 & 94.7 & \textbf{0} & 79.7 & 79.7 & \textbf{0} \\
Gear meshing & 94.8 & 94.8 & \textbf{0} & 96.2 & 96.2 & \textbf{0} & 69.7 & 69.7 & \textbf{0} \\
Nut threading & 89.2 & 89.2 & \textbf{0} & 70.2 & 70.2 & \textbf{0} & 50.6 & 50.6 & \textbf{0} \\
\bottomrule
\end{tabular*}
\end{table}
\end{minipage}\par\medskip

\par\noindent\begin{minipage}{\linewidth}
\setlength{\parskip}{0.5pc}
\subsection{Tests Across Training Seeds}
\label{app:new-seeds}
Table~\ref{tab:new-seeds} summarizes the three-seed test studies. The validation-selected rows use draw 90 for checkpoint selection and draw 100 for testing, with 640 episodes per draw. Training seeds are 1, 2, and 3. The means and sample standard deviations describe variation between these seeds. The matched-machine Full-CAP reruns used in the component ablations are separate evaluations of the selected models. In particular, gear and nut values differ slightly between the two evaluation platforms. For nut CAP-PPO, training uses FORGE rl\_games with an asymmetric critic and the 5\,N CAP shield for 100 epochs (3,276,800 environment steps). Epoch-100 checkpoints are tested on draw 100; draw 90 is used for validation.

\begin{table}[H]
\centering\footnotesize\setlength{\tabcolsep}{5pt}
\caption{\textbf{Tests across training seeds.} Each row aggregates three training seeds and 1,920 test episodes. All force metrics use the filtered decision-rate signal.}
\label{tab:new-seeds}
\begin{tabular}{@{}>{\raggedright\arraybackslash}p{\dimexpr0.1\linewidth-1\tabcolsep\relax}>{\raggedright\arraybackslash}p{\dimexpr0.27\linewidth-2\tabcolsep\relax}>{\raggedleft\arraybackslash}p{\dimexpr0.09\linewidth-2\tabcolsep\relax}>{\raggedleft\arraybackslash}p{\dimexpr0.2\linewidth-2\tabcolsep\relax}>{\raggedleft\arraybackslash}p{\dimexpr0.2\linewidth-2\tabcolsep\relax}>{\raggedleft\arraybackslash}p{\dimexpr0.14\linewidth-1\tabcolsep\relax}@{}}
Task & Method & Cap (N) & SR (\%) & SSR (\%) & Violations\\\midrule
\rowcolor{appgreen}
Peg & CAP-PPO & 5 & 74.6 $\pm$ 0.7 & 74.6 $\pm$ 0.7 & 0/1920 \\
\rowcolor{appgreen}
Peg & CAP-WM & 5 & 97.6 $\pm$ 1.0 & 97.6 $\pm$ 1.0 & 0/1920 \\
\rowcolor{appgreen}
Peg & CAP-WM & 10 & 99.3 $\pm$ 0.2 & 99.3 $\pm$ 0.2 & 0/1920 \\
Peg & TD-MPC2 + depth & 5 & 99.3 $\pm$ 0.1 & 0.6 $\pm$ 0.2 & 1908/1920 \\
Peg & TD-MPC2 + depth & 10 & 99.4 $\pm$ 0.2 & 49.3 $\pm$ 2.1 & 970/1920 \\
\midrule
\rowcolor{appgreen}
Gear & CAP-PPO & 5 & 85.1 $\pm$ 9.5 & 85.1 $\pm$ 9.5 & 0/1920 \\
\rowcolor{appgreen}
Gear & CAP-WM & 5 & 92.4 $\pm$ 0.3 & 92.4 $\pm$ 0.3 & 0/1920 \\
\rowcolor{appgreen}
Gear & CAP-WM & 10 & 93.9 $\pm$ 2.4 & 93.9 $\pm$ 2.4 & 0/1920 \\
Gear & TD-MPC2 + depth & 5 & 95.6 $\pm$ 0.6 & 0.0 $\pm$ 0.0 & 1919/1920 \\
Gear & TD-MPC2 + depth & 10 & 95.9 $\pm$ 0.5 & 12.9 $\pm$ 1.5 & 1662/1920 \\
\midrule
\rowcolor{appgreen}
Nut & CAP-PPO & 5 & 87.6 $\pm$ 3.5 & 87.6 $\pm$ 3.5 & 0/1920 \\
\rowcolor{appgreen}
Nut & CAP-WM & 5 & 89.9 $\pm$ 3.1 & 89.9 $\pm$ 3.1 & 0/1920 \\
\rowcolor{appgreen}
Nut & CAP-WM & 10 & 93.1 $\pm$ 2.2 & 93.1 $\pm$ 2.2 & 0/1920 \\
Nut & TD-MPC2 + depth & 5 & 94.1 $\pm$ 3.4 & 0.0 $\pm$ 0.0 & 1920/1920 \\
Nut & TD-MPC2 + depth & 10 & 94.1 $\pm$ 3.4 & 13.8 $\pm$ 9.8 & 1648/1920 \\
\bottomrule\end{tabular}\end{table}
\end{minipage}\par\medskip

\section{Controller Uncertainty}
\label{app:new-audits}
The peg-insertion uncertainty variant uses one training seed, 32 environments, and 30,000 vector steps with the world-model learning recipe. Validation uses draw 191 with 640 episodes and a 5\,N cap (Table~\ref{tab:new-audits}).

The uncertainty variant intersects controller constraints over a synthetic interval around biased dead-zone estimates. A 5\% fraction of the configured 5\,N dead-zone range gives a 0.25\,N uncertainty radius. Other software controller parameters remain available, and this variant avoids using the future redraw clock. The interval specifies synthetic uncertainty in the dead-zone estimate; the nominal-parameter ablation substitutes fixed nominal values and retains the clock.

\begin{table}[htbp]
\centering\small
\caption{\textbf{Exploratory validation at a 5\,N cap.} Counts refer to episodes with at least one exceedance.}
\label{tab:new-audits}
\begin{tabular}{@{}lrr@{}}\toprule
Variant & SR (\%) & Violations\\\midrule
Bounded dead-zone estimate & 39.5 & 0/640\\\bottomrule
\end{tabular}\end{table}

For the uncertainty variant, the robust interval is empty on 6.54\% of control steps. The audit finds zero worst-case controller-map violations on nonempty steps at a $10^{-4}$\,N tolerance. This evaluates the controller-output bound under the specified uncertainty interval.

\FloatBarrier

\end{document}